 \documentclass[final,3p,twoside,fleqn,sort&compress]{elsarticle}

\usepackage{graphicx}
\usepackage{amssymb}
\usepackage{amsmath}
\usepackage{amsthm}
\usepackage{stmaryrd}

\usepackage{xcolor}

\usepackage{cleveref} 

\usepackage{enumerate}

\usepackage{mathtools,nccmath}

\usepackage{booktabs}

\let\oldfrac\frac
\renewcommand{\frac}[2]{%
  \mathchoice
    {\oldfrac{#1}{#2}}
    {#1/#2}
    {#1/#2}
    {#1/#2}
}

\theoremstyle{plain}
\newtheorem{theorem}{Theorem}[section]
\newtheorem{lemma}[theorem]{Lemma}
\newtheorem{corollary}[theorem]{Corollary}

\newtheorem{proposition}[theorem]{Proposition}
\theoremstyle{definition}

\newtheorem{example}[theorem]{Example}
\theoremstyle{remark}
\newtheorem{remark}[theorem]{Remark}

\newcommand{\lra}{\leftrightarrow}
\renewcommand{\le}{\leqslant}
\renewcommand{\ge}{\geqslant}

\newcommand{\rr}{\backslash}
\newcommand{\lr}{/}

\newcommand{\nateq}[1]{\Tilde{#1}}
\newcommand{\gsub}{\lesssim}
\newcommand{\gequiv}{\approx}
\newcommand{\acut}[2]{\prescript{#1}{}{#2}}
\newcommand{\rre}{\backslash\!\!\backslash}
\newcommand{\lre}{/\!\!/}

\newcommand{\fs}[1]{\mathcal{F}(#1)}
\newcommand{\fr}[1]{\mathcal{R}(#1)}
\newcommand{\fp}[1]{\mathcal{P}(#1)}
\newcommand{\fe}[1]{\mathcal{E}(#1)}
\newcommand{\fn}[1]{\mathbf{#1}}
\newcommand{\bigo}{\mathcal{O}}

\newcommand{\modif}[1]{{\color{black} #1}}
\newcommand{\modiff}[1]{{\color{black} #1}}

\newcommand{\FLTS}{\textbf{FLTS}}

\usepackage{comment}

\usepackage{algorithm}
\usepackage{algpseudocode}
\makeatletter
\def\BState{\State\hskip-\ALG@thistlm}
\makeatother

\algdef{SE}[DOWHILE]{Do}{doWhile}{\algorithmicdo}[1]{\algorithmicwhile\ #1}%

\journal{Information Sciences}

\begin{document}

\begin{frontmatter}


\title{On the solvability of weakly linear systems of fuzzy relation equations\tnoteref{label1}\tnoteref{label2}}
 \tnotetext[label1]{This research was supported by the Science Fund of the Republic of Serbia, GRANT No 7750185, Quantitative Automata Models: Fundamental Problems and Applications - QUAM}
 \tnotetext[label2]{This research was supported by Ministry of Education, Science and Technological Development, Republic of Serbia, Grant No 174013, Contract No. 451-03-68/2022-14/200124}

\makeatletter
\def\@author#1{\g@addto@macro\elsauthors{\normalsize%
    \def\baselinestretch{1}%
    \upshape\authorsep#1\unskip\textsuperscript{%
      \ifx\@fnmark\@empty\else\unskip\sep\@fnmark\let\sep=,\fi
      \ifx\@corref\@empty\else\unskip\sep\@corref\let\sep=,\fi
      }%
    \def\authorsep{\unskip,\space}%
    \global\let\@fnmark\@empty
    \global\let\@corref\@empty  
    \global\let\sep\@empty}%
    \@eadauthor={#1}
}
\makeatother

\author{Stefan Stanimirovi\' c\corref{labcor}}
\ead{stefan.stanimirovic@pmf.edu.rs}
\cortext[labcor]{Corresponding author.}

\author{Ivana Mici\' c}
\ead{ivana.micic@pmf.edu.rs}

\address{University of Ni\v s, Faculty of Sciences and Mathematics, Vi\v segradska 33, 18000 Ni\v s, Serbia}

\begin{abstract}
Systems of fuzzy relation equations and inequalities in which an unknown fuzzy relation is on the one side of the equation or inequality are linear systems. They are the most studied ones, and a vast literature on linear systems focuses on finding solutions and solvability criteria \modiff{for} such systems.
The situation is quite different with the so-called weakly linear systems, in which an unknown fuzzy relation is on both sides of the equation or inequality. Precisely, the scholars have only given the characterization of the set of exact solutions to such systems. This paper describes the set of fuzzy relations that solve weakly linear systems to a certain degree and provides ways to compute them. We pay special attention to developing the algorithms for computing fuzzy preorders and fuzzy equivalences that are solutions to some extent to weakly linear systems. We establish additional properties for the set of such approximate solutions over some particular types of complete residuated lattices. We demonstrate the advantage of this approach via many examples that arise from the problem of aggregation of fuzzy networks.
\end{abstract}


\end{frontmatter}



\allowdisplaybreaks

\section{Introduction}

\subsection{A literature overview}

The systems of fuzzy relation equations and inequalities enjoy many amazing properties and huge application potential. For this reason, the researchers have studied them from different aspects over many decades. The most studied and well-known systems are the ones called \emph{linear systems}. These are the systems in which every inequality or equation take\modiff{s} one of the following forms: $A X \le B$, $B \le A X$, $X A \le B$, $B \le X A$, $AX = B$, or $X A = B$, in which $A$ and $B$ are known fuzzy sets or fuzzy relations, while an unknown fuzzy relation $X$ is on only one side of the inequality or equation. Starting from the pioneering works of Sanchez \cite{S.74,S.76}, linear systems have been a topic of fruitful research from both theoretical and practical aspects. 

So far, the scholars have studied the characterizations and computations of solutions of linear systems over numerous algebraic structures, including
complete lattices \cite{SQ.21,S.18}, complete Brouwerian lattices \cite{S.12,SQWZ.20,QWL.14,QSW.15,WZ.13,W.01,XW.12}, complete residuated lattices \cite{B.15,BB.11,BK.14,DMMT.17,ICST.15,T.20} and BL-algebras \cite{PN.08}. Linear systems have found application\modiff{s} in many fields. Among others, they have applications in describing various \emph{fuzzy networks}, depending on the underlying structure of truth values. For example, linear systems defined over:
\begin{itemize}
    \item \emph{Addition-min structure} describe the quantitative relation of a Peer-to-Peer file-sharing network when we impose the \emph{total} quality demand of download traffic of peers  \cite{LY.12,LW.21,LY.20,Y.14,Y.21,YLZC.18,Y.18,YZC.15b,Y.20b},
    \item \emph{Max-min fuzzy structure} describe the quantitative relation of the same fuzzy network when we impose the \emph{highest} quality demand of download traffic of peers 
    \cite{YYH.17},
    \item \emph{Max-product structure} describe wireless connected Server-to-Client (S2C) fuzzy network 
    \cite{Y.21.tfs}, wireless communication station system \cite{QLY.21,YZC.15,YZC.16,YYC.18}, foodstuff supply \cite{A.10,A.14,A.12}, various decision-making problems \cite{M.04},
    \item \emph{Min-product structure} describe the supply chain fuzzy network when we impose the price requirements \cite{Y.20}.
\end{itemize}

However, linear systems may not always be solvable. Because of their application potential, it became clear in the early stages of their investigation that it is unreasonable to disregard fuzzy relations that are “close enough'' solutions to these systems. For this purpose, numerous techniques for finding such solutions have been developed in many papers \cite{BV.05,CL.97,G.94,LL.10,P.04,P.13,PN.07,WLWL.21,XZCY.19,Y.19}.

Besides linear systems, several researchers also considered the \emph{weakly linear systems} of fuzzy relation equations and inequalities, in which every inequality or equation takes one of the following forms: $A X \le X A$, $X A \le A X$, or $A X = X A$, in which $A$ is a known fuzzy relation, while an unknown fuzzy relation $X$ is on both sides of the inequality or equation. Various researchers have studied such systems and many generalized versions of such systems over various algebraic structures. Let us enumerate complete residuated lattices \cite{IC.12,ICB.10,ICDJ.12,SCI.17,QZ.20} and max-plus algebras \cite{SCD.21}, among others. \modif{Such structures are suitable for studying weakly linear systems because there exists the greatest solution to these systems. Moreover, the authors of the mentioned papers have modified the Kleene Fixed Point Theorem to obtain iterative methods to compute this greatest solution. However, the drawback of this methodology is that the proposed methods do not necessarily finish in a finite number of iterations. Due to this drawback, the authors have studied sufficient conditions for the termination of their methods.} \modif{It should be noted that weakly linear} systems have arisen from the well-studied problems in fuzzy automata theory, including state reduction of fuzzy automata (see \cite{CIDB.12,CIJD.12,CSIP.07,CSIP.10,MJS.18,SCI.14}). Simultaneously, the positional analysis and blockmodeling of fuzzy social networks solve these systems. Specifically, Fan et al. \cite{FL.14,FLL.08,FLL.07} have defined \emph{regular fuzzy equivalences} exactly as those solutions to weakly linear systems that are fuzzy equivalences. They are able to model how the nodes are similar to each other. We can use these fuzzy equivalences to construct the “aggregated'' fuzzy network by grouping the similar nodes of the original fuzzy network (see \cite{CB.10,SCI.17} for the way of constructing such a fuzzy network).

\subsection{Motivation and argumentation}

Weakly linear systems, as defined above, are always solvable. Indeed, such systems always have the \emph{trivial solution} for every fuzzy relation $A$, the identity relation $I$. However, the trivial solution is irrelevant in applications because the “aggregated'' fuzzy network is the same as the staring fuzzy network. Nevertheless, it is not uncommon that only the trivial solution exists for such systems.

Besides, even when a nontrivial solution exists, it can happen that the well-known procedures for computing such a solution do not terminate. Such a situation may happen when the underlying lattice is not locally finite (e.g., the max-product structure). Ignjatovi\' c et al. \cite{IC.12,ICB.10,ICDJ.12} have already proposed some alternatives in such cases, including finding crisp solutions or imposing additional restrictions on the lattice.

In this research, we argue that, as for linear systems, it is unreasonable to disregard fuzzy relations that are “close enough'' solutions to weakly linear systems since they behave better in previous cases than the proposed alternatives. The study of such solutions has multiple advantages. First, when a weakly linear system has only the trivial solution, one can find a nontrivial approximate solution. Also, consider a situation when there is an exact nontrivial solution\modiff{,} and the known procedures for its computation do not terminate. To overcome this problem, we may pick a fuzzy relation that is a solution to a sufficiently high degree but for which the procedure for its computation terminates.

As we show in this paper, the greatest solution to a certain degree to any weakly linear system always exists, regardless of a chosen degree. However, we are not interested in arbitrary such fuzzy relations for the reduction of fuzzy networks, but in those that are fuzzy equivalences or fuzzy preorders. This is because we can group the nodes of a fuzzy network according to classes of fuzzy equivalences when we use a fuzzy equivalence, or according to aftersets or foresets of fuzzy preorders when we use a fuzzy preorder. However, the structure of the set of all fuzzy preorders (fuzzy equivalences), which are solutions to a certain degree to weakly linear systems, is entirely different compared to the one in a non-approximate case. Namely, as shown in \cite{ICB.10}, for a given fuzzy preorder (fuzzy equivalence)\modiff{,} there always exists the greatest fuzzy preorder (fuzzy equivalence) contained in it. On the other hand, there may not exist the greatest fuzzy preorder (fuzzy equivalence) that is a solution to a desirable degree, and that is contained in a given fuzzy preorder (fuzzy equivalence). This holds only for some particular types of complete residuated lattices. The additional problem is that the known methodologies cannot even find a maximal fuzzy preorder that is a solution to some degree. Nonetheless, we show that we can still substantially reduce fuzzy networks even when our procedure does not return the greatest or a maximal fuzzy preorder that is a solution to the desirable degree. This is of particular interest for fuzzy networks that cannot be reduced using fuzzy preorders (fuzzy equivalences) that are the exact solutions.

\subsection{The structure of the paper}

Section \ref{sec.intro} evokes necessary notions and notations regarding complete residuated lattices (which serve as a structure of truth values), fuzzy sets, and fuzzy relations. 
In Section \ref{sec.SD}, we introduce the notions that allow us to study the solvability degree of weakly linear systems (WLSs in what follows) and examine the set of fuzzy relations that solve WLSs to a certain degree. Further, we identify additional properties of WLSs defined over the Heyting algebra. We develop a procedure for computing the greatest fuzzy relation that solves a WLS to a certain degree in Section \ref{sec.CGAS}. 
Likewise, in Section \ref{sec.CGFP}, we propose an adaptation of the previous algorithm according to which it computes a fuzzy preorder (or a fuzzy equivalence). In the end, we show the applications of the developed methodology in the aggregation of fuzzy networks in Section \ref{sec.App}.

\section{Preliminaries} \label{sec.intro}

%

Recall that a lattice can be defined either as a partially ordered set $(L, \le)$ in which supremum and infimum exist for every finite subset of $L$ or as an algebraic structure $(L, \land, \lor)$ in which binary operations $\land$ and $\lor$ on $L$ satisfy the laws of commutativity, associativity, and absorption. These two definitions are mutually equivalent. In addition, a lattice is complete if supremum and infimum exist for an arbitrary subset of $L$. If we have an operation $\otimes$ on a lattice $L$ (called \emph{multiplication}) such that the structure $(L, \otimes, 1)$ is a commutative monoid, then the unique operation $\to$ on $L$ that satisfies the {\it residuation property\/}:
\begin{equation}\label{eq:adj}
x\otimes y \leqslant z \ \textrm{ iff } \ x\leqslant y\to z, \quad \textrm{ for each }x, y, z\in L.
\end{equation}
is called the \emph{residuum}. Moreover, if $(L, \land, \lor, 0, 1)$ is a (complete) bounded distributive lattice, then the structure ${\cal L}=(L, \land, \lor, \otimes, \to, 0, 1)$ is a \emph{(complete) residuated lattice}. 
%
In such structure, we can naturally define the \emph{biresiduum} (or \emph{biimplication}) as $x \lra y = (x \to y) \land (y \to x)$, for every $x, y \in L$. Note that $\otimes$ preserves the order in both arguments, while $\to$ preserves the order only in the second argument. That means that $\to$ reverses the order in the first argument. 

For \modif{scalars $x$, $y$, $z$} and an indexed family of scalars $\{y_i\}_{i \in I}$ from a complete residuated lattice $\cal L$, the following holds (see \cite{B.02a,BV.05} for more details):
\begin{align}
    &x \otimes (x \to y) \le y, \\
    &(x \otimes y) \to z = x \to (y \to z), \label{eq.CRLProp1}\\
    &x \otimes \left( \bigvee_{i \in I} y_i \right) = \bigvee_{i \in I} (x \otimes y_i),\label{eq.LandSup} \\
    &\bigvee_{i \in I} x_i \to y = \bigwedge_{i \in I} (x_i \to y), \label{eq.CRLProp2}\\
    &x \to \bigwedge_{i \in I} y_i = \bigwedge_{i \in I} (x \to y_i), \label{eq.CRLProp3}\\
    &x \otimes \left( \bigwedge_{i \in I} y_i \right) \le \bigwedge_{i \in I} (x \otimes y_i). 
\end{align}

An instance of a residuated lattice is $L=[0, 1]$ equipped with $x\land y =\min
(x,y)$ and $x\lor y =\max (x,y)$. On the real-unit interval we can define three structures of truth values: the {\it product} ({\it Goguen\/}) {\it structure\/}, the {\it G\"odel structure\/} and the {\it {\L}ukasiewicz structure\/},
where $\otimes$, $\to$ and $\lra$ are defined with $x\otimes y = x\cdot y$, $x\to y= 1$ if $x\le y$ and~$=y/x$ otherwise, $x \lra y = 1$ if $x=y=0$ and $=\min(x,y)/\max(x,y)$ otherwise for the product structure, $x\otimes y = \min(x,y)$, $x\to y= 1$ if $x\le y$ and $=y$ otherwise, $x \lra y = 1$ if $x=y$ and $=\min(x,y)$ otherwise for the G\"odel structure, as well as $x\otimes y = \max(x+y-1,0)$, $x\to y=\min(1-x+y,1)$ and $x \lra y = 1 - |x - y|$ for the {\L}ukasiewicz structure.
A residuated lattice $\cal L$ is \emph{locally finite} if any finitely generated subalgebra of $\cal L$ is finite. 
The G\" odel structure is an example of a locally finite lattice, while the product structure is an example of an opposite one.

A residuated lattice is \emph{idempotent} if the operation $\otimes$ is idempotent, or in other words, if $x \otimes x = x$ for every $x \in L$. Note that $\otimes$ is idempotent if and only if $\otimes = \land$. Indeed, if $\otimes$ is idempotent, then $x \land y = (x \land y) \otimes (x \land y) \le x \otimes y$, whereas the other inequality holds in every residuated lattice (cf.~\cite{B.02a,BV.05}). The other direction follows immediately. A residuated lattice in which $\otimes = \land$ is also known as a \emph{Heyting algebra} (cf.~\cite{E.19,R.07}). According to this, a residuated lattice is idempotent if and only if it is a Heyting algebra.

For a given complete residuated lattice $\cal L$ and a nonempty set $U$ (the \emph{universe}), we define a \emph{fuzzy set over $\cal L$ and $U$}, or just a \emph{fuzzy set}, as any mapping from $U$ into $L$. By $\fs{U}$ we denote the set of all fuzzy sets over $\cal L$ and $U$. The set $\fs{U}$ is equipped with the equality and the inclusion (ordering) of fuzzy sets (defined as for the ordinary mappings), as well with the meet (intersection) and the join (union) defined coordinate-wise. A crisp subset of $U$ is a fuzzy set over $\{0,1\}\subseteq L$ and $U$. The \emph{$x$-cut} of a fuzzy set $A \in \fs{U}$ is a crisp subset $\acut{x}{A}$ given by $\acut{x}{A} = \{ u \in U | A(u) \ge x \}$, for every $x \in L$.

A {\it binary fuzzy relation on $U$\/}, or just a \emph{fuzzy relation on $U$} in what follows, is any mapping from $U\times U$ to $L$. The set $\fr{U}$ of all fuzzy relations on $U$ is equipped with the equality, ordering, meet and join. A crisp relation is a fuzzy relation over the set $\{0, 1\} \subseteq L$. 
The \emph{inverse} of a fuzzy relation $R$ on $U$ is a fuzzy relation $R^{-1}$ on $U$ defined with $R^{-1}(v,u) = R(u,v)$, for every $u, v \in U$. 
A particularly important fuzzy relation used thought the rest of the paper is the \emph{identity relation} on $U$, denoted by $\vartriangle_{U}$ and defined by $\vartriangle_{U}(u, u) =1$ and $\vartriangle_{U}(u, v) =0$, for $u, v \in U$ such that $u \neq v$. In addition, $\blacktriangle_U$ is the \emph{universal relation} on $U$, defined by $\blacktriangle_U(u, v) = 1$ for every $u, v \in U$.

The {\it composition\/} of two fuzzy relations $R, P \in \fr{U}$ is a fuzzy relation $R\circ P \in \fr{U}$ defined for every $u,v \in U$ by:
\begin{equation}\label{eq.fuzzyRelComp}
    (R \circ P )(u,v) = \bigvee_{w\in U}\,R(u,w)\otimes P(w,v).
\end{equation}
The \emph{$n$th degree} of a fuzzy relation $R \in \fr{U}$ is inductively defined as $R^0 = \vartriangle_U$, $R^1 = R$ and $R^{n + 1} = R \circ R^{n}$, for every $n \in \mathbb{N}_0$.
The composition of fuzzy relations is associative and preserves order in both arguments.
Moreover, for $R, P, Q \in \fr{U}$, $R_i\in \fr{U} (i \in I)$ and $x \in L$, the following holds:
\begin{align}
    & P \circ \left( \bigvee_{i \in I} R_i \right)
    = \bigvee_{i \in I} (P \circ R_i),  \ 
    \left( \bigvee_{i \in I} R_i \right) \circ P
    = \bigvee_{i \in I} (R_i \circ P), \label{eq.CircBigVee}
\end{align}

For a scalar $x \in L$ and a fuzzy relation $R \in \fr{U}$, fuzzy relations $x \otimes R \in \fr{U}$ and $x \to R \in \fr{U}$ are defined as:
\begin{align}
    & (x \otimes R)(v,w) = x \otimes R(v,w), \label{eq.fuzzyRelOtimes}\\
    & (x \to R)(v,w) = x \to R(v,w), \label{eq.fuzzyRelTo}
\end{align}
for every $v,w \in U$. In the sequel, we suppose that $\circ$, defined by \eqref{eq.fuzzyRelComp}, has a higher precedence than $\otimes$ and $\to$ defined by \eqref{eq.fuzzyRelOtimes} and \eqref{eq.fuzzyRelTo}.

We define two fuzzy relations $\gsub$ and $\gequiv$ over $\fs{U}$ and $\cal L$ for every $A, B  \in \fs{U}$ as:
\begin{align}
    & \gsub(A, B) = \bigwedge_{u \in U} A(u) \to B(u), \\
    & \gequiv(A, B) = \bigwedge_{u \in U} A(u) \lra B(u).
\end{align}
We use infix notations in what follows, and write $A \gsub B$ rather than $\gsub(A, B)$, as well as $A \gequiv B$ rather than $\gequiv(A, B)$. Intuitively, $A \gsub B$ measures the \emph{inclusion degree} of the fuzzy set $A$ in the fuzzy set $B$. Similarly, $A \gequiv B$ measures the \emph{degree of equality} of fuzzy sets $A$ and $B$. The following properties that hold for every $A, B, C, D \in \fs{U}$ are used through the rest of the paper (see \cite{B.02a} for more details):
\begin{align}
    & A \gsub A = 1, \quad A \gequiv A = 1, \label{eq.gsubgeq1}\\
    & A \gequiv B = B \gequiv A,  \label{eq.gsubgeq2}\\
    & A \gequiv B = (A \gsub B) \land (B \gsub A), \label{eq.gsubgeq3} \\
    & A \gsub B = 1 \ \textrm{ iff }\  A \le B, \quad
    A \gequiv B = 1 \ \textrm{ iff }\  A = B, \label{eq.gsubgeq4}\\
    & (A \gsub B) \otimes (B \gsub C) \le A \gsub C, \label{eq.gsubgeq5}\\
    & (A \gequiv B) \otimes (B \gequiv C) \le A \gequiv C, \\
    & (A \gequiv B) \otimes (B \gsub C) \otimes (C \gequiv D ) \le  A \gsub D,\label{eq.gsubgeq6}
\end{align}
Moreover, for every $A, A_i, B, B_i \in \fs{U}$, where $i \in I$, we have:
\begin{align}
    & A \gsub \left( \bigwedge_{i \in I} B_i \right) = \bigwedge_{i \in I} A \gsub B_i,\label{eq.gsubgeq9} \\
    & \left( \bigvee_{i \in I} A_i \right) \gsub B = \bigwedge_{i \in I} A_i \gsub B. \label{eq.gsubgeq10}
\end{align}
In addition, for fuzzy relations $R_1, R_2, P_1, P_2 \in \fr{U}$, we have:
\begin{align}
    & (R_1 \gsub R_2) \otimes (P_1 \gsub P_2) \le (R_1 \circ P_1) \gsub (R_2 \circ P_2), \label{eq.gsubgeq7}\\
    & (R_1 \gequiv R_2) \otimes (P_1 \gequiv P_2) \le (R_1 \circ P_1) \gequiv (R_2 \circ P_2).\label{eq.gsubgeq8}
\end{align}
In the following lemma, we prove a simple result which is needed for the rest of the work.
\begin{lemma}
When $A, B \in \fs{U}$, then:
\begin{align} 
    & A \otimes (A \gsub B) \le B, \label{eq.AGsubBOtALeB} \\
    & A \otimes (A \gequiv B) \le B. \label{eq.AGequBOtALeB}
\end{align}
\end{lemma}
\begin{proof}
Choose an arbitrary $w \in U$. Then the following is valid:
\begin{align*}
    A(w) \otimes (A \gsub B) = 
    A(w) \otimes \bigwedge_{u \in U} (A(u) \to B(u)) \le A(w) \otimes (A(w) \to B(w)) \le B(w),
\end{align*}
which means that \eqref{eq.AGsubBOtALeB} follows. In addition, \eqref{eq.AGequBOtALeB} immediately follows since $A \gequiv B \le A \gsub B$.
\end{proof}

A fuzzy relation $P \in \fr{U}$ is \emph{reflexive} if $\vartriangle_{U} \le P$, \emph{symmetric} if $P^{-1} \le P$, and \emph{transitive} if $P \circ P \le P$. A \emph{fuzzy preorder} is a reflexive and transitive fuzzy relation.
The set $\fp{U}$ of all fuzzy preorders on $U$ is a complete lattice such that the meet is the same as in the lattice $\fr{U}$, but the joins in these two lattices do not necessarily coincide. 
Evidently, if $P$ is a fuzzy preorder, then $P \circ P = P$. 
Moreover, a \emph{fuzzy equivalence} is a symmetric fuzzy preorder. Likewise, as for the set $\fp{U}$, the set $\fe{U}$ of all fuzzy equivalences on a set $A$ is also a complete lattice. This lattice shares only the meet with the lattice $\fr{U}$.

Let $P \in \fr{U}$ and $u \in U$. The \emph{$P$-afterset of $u$} is the fuzzy subset $u P \in \fs{U}$ defined by $(uR)(v) = P(u, v)$, for every $v \in U$. Dually, the \emph{$P$-foreset of $u$} is the fuzzy subset $P u \in \fs{U}$ defined by $(P u)(v) = P(v, u)$, for every $v \in U$. The set of all $P$-aftersets (resp. all $P$-foresets) is denoted by $U P$ (resp. $P U$). If $P \in \fe{U}$, then for every $u \in U$ we have that $u P = P u$, and this fuzzy subset is the \emph{equivalence class of $P$ determined by $u$} (cf. \cite{CIB.07}). The equivalence class of $P$ determined by $u$ is as usual denoted by $P_{u}$, and the set of all equivalence classes of $P$ by $U/P$.

There exists a natural link between fuzzy preorders and fuzzy equivalences. Indeed, if $P \in \fp{U}$ is a fuzzy preorder on $U$, then a fuzzy relation $\tilde{P} = P \wedge P^{-1} $ is a fuzzy equivalence on $U$, and is called the \emph{natural fuzzy equivalence} of $P$. It is well-known that if two elements $u, v \in U$ are connected in the natural fuzzy equivalence in the degree $1$, then it is equivalent to say that equivalence classes $\tilde{P}_{u}$ and $\tilde{P}_{v}$ are equal. Or, equivalently, to say that the aftersets $u P$ and  $v P$ (resp. foresets $P u$ and $P v$) are equal. Moreover, the number of all equivalence classes of $\tilde{P}$ is the same as the number of aftersets of $P$ (and further the same as the number of foresets of $P$, cf. \cite{CIB.07}).

Although the composition of fuzzy relations is associative, it is not commutative in general. Thus, if we consider the composition of fuzzy relations as an operation of multiplication of fuzzy relations, then there exists two operations that satisfy the relationships analogous to the residuation property~\eqref{eq:adj}. More precisely, if $R, Q \in \fr{U}$ are fuzzy relations and $X \in \fr{U}$ is an unknown fuzzy relation, then there exists the greatest solution to the inequality $R \circ X \le Q$ in the set $\fr{U}$, called the \emph{right residual of $Q$ by $R$}, which is denoted by $R \rr Q \in \fr{U}$. Dually, there exists the greatest solution of the inequality $X \circ R \le Q$ in the set $\fr{U}$, called the \emph{left residual of $Q$ by $R$}, and is denoted by $Q \lr R \in \fr{U}$. These two residuation properties can be written as:
\begin{align}
    &R \circ X \le Q \quad \textrm{ iff } \quad X \le R \rr Q, \label{eq.rightAdjFuzzyRel}\\
    &X \circ R \le Q \quad \textrm{ iff } \quad X \le Q \lr R. \label{eq.leftAdjFuzzyRel}
\end{align}
It can be easily verified that $R \rr Q$ and $Q \lr R$ can be calculated for every $u, v \in U$ in the following way:
\begin{align*}
    & (R \rr Q)(u,v) = R u \gsub Q v \\
    & (Q \lr R)(u,v) = v R \gsub u Q.
\end{align*}
For more information on right and left residuals of fuzzy relations, we refer to \cite{IC.12,ICB.10,ICDJ.12}. Moreover, we extend the previous notations and define the following fuzzy relations on a set $U$:
\begin{align*}
    & R \rre Q = (R \rr Q) \land (Q \rr R)^{-1},\\
    & Q \lre R = (Q \lr R) \land (R \lr Q)^{-1}.
\end{align*}
It can be shown that for each $u, v \in U$ we have:
\begin{align*}
    & (R \rre Q)(u,v) = R u \gequiv Q v \\
    & (Q \lre R)(u,v) = v R \gequiv u Q.
\end{align*}

We use the following two lemmas \modiff{throughout} the rest of the paper (see\modiff{,} for example\modiff{,} \cite{ICB.10}).

\begin{lemma} \label{lem.AuxFuzzyRel1}
Let $R \in \fr{U}$. Then $R \lr R \in \fp{U}$ and $R \rr R \in \fp{U}$.
\end{lemma}

\begin{lemma} \label{lem.AuxFuzzyRel2}
Let $R_1, R_2 \in \fp{U}$. Then $R_1 \land R_2 \in \fp{U}$.
\end{lemma}

\section{The solvability degree of weakly linear systems} 
\label{sec.SD}

We assume that $U$ is a universe set and $\fn{S} = \{ R_i\}_{i \in I}$ an indexed family of fuzzy relations on $U$. The following systems of inequalities and equations:
\begin{align}
    & X \circ  R_i \le  R_i \circ X, \label{eq.WLS1}\\
    &  R_i \circ X \le X \circ  R_i, \label{eq.WLS2}\\
    & X \circ  R_i =  R_i \circ X,  \label{eq.WLS3}
\end{align}
for every $i \in I$,
in which $X \in \fr{U}$ is an unknown fuzzy relation on $U$, are called \emph{weakly linear systems} (WLSs, for short). \modif{In what follows, we also use abbreviations WLS-1, WLS-2 and WLS-3 for systems \eqref{eq.WLS1}, \eqref{eq.WLS2} and \eqref{eq.WLS3}, respectively.}

With this in mind, for a given fuzzy relation $ R \in \fr{U}$, we introduce three fuzzy subsets $SD_1( R), SD_2( R)$ and $SD_3( R)$ on the set of all fuzzy relations $\fr{U}$ as:
\begin{align}
    & [SD_1( R)](X) = X \circ  R \gsub  R \circ X, \\
    & [SD_2( R)](X) = R \circ X \gsub X \circ  R, \\
    & [SD_3( R)](X) = X \circ  R \gequiv  R \circ X,
\end{align}
for every $X \in \fr{U}$. It comes straightforward from \eqref{eq.gsubgeq3} that $SD_3( R) = SD_1( R) \land SD_2( R)$. Intuitively, the value $[SD_1( R)](X)$ (resp., $[SD_2( R)](X)$) measures the degree to which $X$ solves the inequality $X \circ  R \le  R \circ X$ (resp., $ R \circ X \le X \circ  R$). Likewise, the value $[SD_3( R)](X)$ measures the degree to which $X$ solves the equation $X \circ  R =  R \circ X$. Thus, we call the value $[SD_k(R)](X)$ the \emph{solution degree} for every $X$ and $k \in \{1, 2, 3\}$.

Consequently, for a given family $\fn{S} = \{ R_i\}_{i \in I}$ of fuzzy relations on $U$, we define fuzzy subsets $SD_k(\fn{S})$, for every $k \in \{1, 2, 3\}$, on the set $\fr{U}$ as:
\begin{equation}
    [SD_k(\fn{S})](X) = \bigwedge_{i \in I} [SD_k( R_i)](X),
\end{equation}
for every $X \in \fr{U}$. For every $k \in \{1,2,3\}$, the value $[SD_k(\fn{S})](X)$ models the degree to which $X$ is a solution to WLS-$k$.

For every $x \in L$ and $k \in \{1, 2, 3\}$, the set $\acut{x}{SD_k(\fn{S})}$ consists of all fuzzy relations that are solutions to the corresponding WLS in a degree at least the chosen degree $x$ from the complete residuated lattice $\cal L$. Of course, fuzzy relations belonging to the set $\acut{1}{SD_k(\fn{S})}$ are the \emph{solutions} to WLSs.

In what follows, we are usually not interested only in fuzzy relations which are solutions to WLSs to a certain degree, but in such fuzzy relations contained in a given fuzzy relation $X_0 \in \fr{U}$. This is also very important when designing the algorithms for computing such fuzzy relations. With $\acut{x}{SD_k(\fn{S}, X_0)}$ we denote the set $\acut{x}{SD_k(\fn{S})} \cap \{ X \in \fr{U} | X \le X_0\}$ of all fuzzy relations that solve WLS-$k$ to the degree at least $x$ and which are included in some fixed fuzzy relation $X_0$, for every $k \in \{1,2,3\}$. As we show in this section, this slight complication in notation does not affect the structure of the set $\acut{x}{SD_k(\fn{S})}$, or in other words, all the properties that hold in $\acut{x}{SD_k(\fn{S})}$ also hold in $\acut{x}{SD_k(\fn{S}, X_0)}$.

\begin{remark}
Through the rest of this paper, if not stated otherwise, $\fn{S} = \{ R_i\}_{i \in I}$ denotes a family of fuzzy relations on a universe $U$, and $X_0$ denotes some fuzzy relation on $U$.
\end{remark}

\modiff{With the following Lemma, we show that we increase the set of all approximate solutions to WLS by decreasing the solution degree.}

\begin{lemma}
Pick two scalars $x$ and $y$ from the complete residuated lattice $L$ so that $x \le y$ holds. Then $\acut{y}{SD_k(\fn{S}, X_0)} \subseteq \acut{x}{SD_k(\fn{S}, X_0)}$.
\end{lemma}
\begin{proof}
Let $X \in \acut{y}{SD_1(\fn{S}, X_0)}$. This means that $y \le (X \circ  R_i \gsub  R_i \circ X)$, for every $i \in I$. But from the assumption, we obtain  $x \le (X \circ  R_i \gsub  R_i \circ X)$, for every $i \in I$. Therefore,  $X \in \acut{x}{SD_1(\fn{S}, X_0)}$. We prove the cases when $k=2$ and $k=3$ similarly.
\end{proof}

In order to give the characterization of the set $\acut{x}{SD_k(\fn{S},X_0)}$, for every $x \in L$ and $k \in \{1, 2, 3\}$, we first prove next results.

\begin{lemma} \label{lem.supSolutACutAux}
For every $k \in \{1, 2, 3\}$ the following holds:
\begin{equation}\label{eq.SDkSup}
    \bigwedge_{j \in J} [SD_k(\fn{S},X_0)](X_j) \le [SD_k(\fn{S},X_0)]\bigg(\bigvee_{j \in J} X_j\bigg). 
\end{equation}
\end{lemma}
\begin{proof}
We prove only the case when $k=1$. Indeed, note that $X_m \le \bigvee_{j \in J} X_j$, for every $m \in J$. Thus, from \eqref{eq.gsubgeq4} we have $X_m \gsub \bigg(\bigvee_{j \in J} X_j\bigg) = 1$, for every $m \in J$. Further, from \eqref{eq.gsubgeq7} we have that for every $m \in J$:
\begin{align*}
    1 & = X_m \gsub \bigg(\bigvee_{j \in J} X_j\bigg) = 1 \otimes \Bigg(X_m \gsub \bigg(\bigvee_{j \in J} X_j\bigg)\Bigg)
    = ( R_i \gsub  R_i) \otimes \Bigg(X_m \gsub \bigg(\bigvee_{j \in J} X_j\bigg)\Bigg)\\
    & \le ( R_i \circ X_m) \gsub \Bigg( R_i \circ \bigg(\bigvee_{j \in J} X_j\bigg)\Bigg).
\end{align*}
This means that we can put the sign $=$ instead of $\le$ in the last line, which further yields that for every $m \in J$ and $i \in I$:
\begin{align*}
    [SD_1( R_i)](X_m) & = (X_m \circ  R_i \gsub  R_i \circ X_m) \otimes 1 \\
    & = (X_m \circ  R_i \gsub  R_i \circ X_m) \otimes 
    \Bigg(  R_i \circ X_m \gsub  R_i \circ \bigg(\bigvee_{j \in J} X_j\bigg)\Bigg) \\
    & \le \Bigg(X_m \circ  R_i \gsub  R_i \circ \bigg(\bigvee_{j \in J} X_j\bigg)\Bigg).
\end{align*}
Finally, \eqref{eq.gsubgeq10} and \eqref{eq.CircBigVee} yield that for every $i \in I$:
\begin{align*}
    \bigwedge_{m \in J} [SD_k( R_i)](X_m) &\le \bigwedge_{m \in J} \Bigg(X_m \circ  R_i \gsub  R_i \circ \bigg(\bigvee_{j \in J} X_j\bigg)\Bigg) \\
    & = \Bigg(\bigg(\bigvee_{m \in J} X_m \circ  R_i\bigg) \gsub  R_i \circ \bigg(\bigvee_{j \in J} X_j\bigg)\Bigg)\\
    & = \Bigg(\bigg(\bigvee_{m \in J} X_m \bigg) \circ  R_i \gsub  R_i \circ \bigg(\bigvee_{j \in J} X_j\bigg)\Bigg)\\
    & = \Bigg(\bigg(\bigvee_{j \in J} X_j \bigg) \circ  R_i \gsub  R_i \circ \bigg(\bigvee_{j \in J} X_j\bigg)\Bigg)\\
    &= [SD_1( R_i)]\bigg(\bigvee_{j \in J} X_j\bigg),
\end{align*}
which means that for every $i \in I$:
\[
\bigwedge_{j \in J} [SD_k( R_i)](X_j) \le [SD_1( R_i)]\bigg(\bigvee_{j \in J} X_j\bigg).
\] 
Infimum over $i \in I$ can go through the previous inequality, thus the statement of the Lemma follows when $k = 1$. Other cases can be proved similarly.
\end{proof}

\begin{lemma} \label{lem.supSolutACut}
Let $k \in \{1, 2, 3\}$ and $x \in L$. Then, if $X_j \in \acut{x}{SD_k(\fn{S},X_0)}$, for every $j \in J$, then 
\begin{equation} \label{eq.varphijXcutSDk}
    \bigvee_{j \in J}X_j \in \acut{x}{SD_k(\fn{S},X_0)}.
\end{equation}
In other words, $\acut{x}{SD_k(\fn{S},X_0)}$ is a complete join-semilattice.
\end{lemma}
\begin{proof}
If $X_j \in \acut{x}{SD_k(\fn{S},X_0)}$ for every $j \in J$, then $x \le [SD_k(\fn{S},X_0)](X_j)$  for every $j \in J$. That means that
\[
x \le \bigwedge_{j \in J} [SD_k(\fn{S},X_0)](X_j).
\]
Lemma \ref{lem.supSolutACutAux} further yields that
\[
x \le [SD_k(\fn{S},X_0)]\bigg(\bigvee_{j \in J}X_j\bigg),
\]
which means that \eqref{eq.varphijXcutSDk} indeed holds.
\end{proof}

\begin{theorem} \label{thm.GreatestLambdaRegRel}
For every $x \in L$ and $k \in \{1, 2, 3\}$, the set $\acut{x}{SD_k(\fn{S},X_0)}$ is a complete lattice.
\end{theorem}
\begin{proof}
Note that $\emptyset \in \acut{x}{SD_k(\fn{S},X_0)}$, which means that $\acut{x}{SD_k(\fn{S},X_0)}$ is nonempty. Denote with $X$ the join of all elements from the set $\acut{x}{SD_k(\fn{S},X_0)}$. According to Lemma~\ref{lem.supSolutACut}, $X\in \acut{x}{SD_k(\fn{S},X_0)}$, and by the construction, $X$ is the greatest element of $\acut{x}{SD_k(\fn{S},X_0)}$. This means that $\acut{x}{SD_k(\fn{S},X_0)}$ is a complete lattice where $X$ is the greatest element of this set.
\end{proof}

\modiff{In what follows, we prove that when we restrict ourselves to complete residuated lattices with $\otimes$ being idempotent, we can prove additional properties for fuzzy relations that are solutions to WLSs to some degree.} Recall that in such structures, $\otimes = \land$ is satisfied, i.e., such complete residuated lattices are actually complete Heyting algebras.

First, we prove that if $X$ is a solution to WLS to some degree, and $X'$ is equal to $X$ to some degree, then $X'$ is also a solution to WLS to a degree at least the conjunction of the previous two degrees.

\begin{lemma} \label{lem.FirstLemaApproxSol}
If $X, X' \in \fr{U}$ are fuzzy relations defined over a Heyting algebra $\cal L$, then for every $k \in \{1, 2, 3\}$ we have:
\begin{equation}
    [SD_k(\fn{S},X_0)](X) \land (X \gequiv X') =  [SD_k(\fn{S},X_0)](X') \land (X \gequiv X').
\end{equation}
\end{lemma}
\begin{proof}
We give the proof only in the case $k=1$, because the proof in other cases follows similarly. 
First, we want to prove that the following holds:
\begin{equation}\label{eq.FirstLemaApproxSol}
    [SD_k(\fn{S},X_0)](X) \land (X \gequiv X') \le [SD_k(\fn{S},X_0)](X').
\end{equation}
Indeed, starting from the left-hand side of \eqref{eq.FirstLemaApproxSol}, we obtain
\begin{align*}
    [SD_k(\fn{S},X_0)](X) \land (X \gequiv X')
    &= \bigwedge_{i \in I} (X \circ  R_i \gsub  R_i \circ X) \land (X \gequiv X')\\
    &\le \bigwedge_{i \in I} [(X \circ  R_i \gsub  R_i \circ X) \land (X \gequiv X')].
\end{align*}
Using the fact that $\land$ is idempotent and commutative, as well that \eqref{eq.gsubgeq1} and \eqref{eq.gsubgeq8} hold, we obtain
\begin{align*}
    &[SD_k(\fn{S},X_0)](X) \land (X \gequiv X')\\
    & \le \bigwedge_{i \in I} [(X \circ  R_i \gsub  R_i \circ X) \land ( R_i \gequiv  R_i) \land (X \gequiv X') \land (X \gequiv X') \land ( R_i \gequiv  R_i)] \\
    & \le \bigwedge_{i \in I} [(X \circ  R_i \gsub  R_i \circ X) \land ( R_i \circ X \gequiv  R_i \circ X') \land
    (X \circ  R_i \gequiv X' \circ  R_i)].
\end{align*}
Finally, by employing \eqref{eq.gsubgeq2} and \eqref{eq.gsubgeq6}, we get
\begin{align*}
    [SD_k(\fn{S},X_0)](X) \land (X \gequiv X')
    \le \bigwedge_{i \in I} (X' \circ  R_i \gsub  R_i \circ X')  = [SD_k(\fn{S},X_0)](X'),
\end{align*}
which proves \eqref{eq.FirstLemaApproxSol}. But, \eqref{eq.FirstLemaApproxSol} also implies that
\begin{equation}\label{eq.FirstLemaApproxSolTemppp}
    [SD_k(\fn{S},X_0)](X') \land (X' \gequiv X) \le [SD_k(\fn{S},X_0)](X).
\end{equation}
Again, since \eqref{eq.gsubgeq2} holds, and $\land$ is idempotent, the statement of the Lemma follows by multiplying both \eqref{eq.FirstLemaApproxSol} and \eqref{eq.FirstLemaApproxSolTemppp} with $(X \gequiv X')$.
\end{proof}

As a direct consequence of Lemma~\ref{lem.FirstLemaApproxSol}, we get that, in linearly ordered Heyting algebras, the degree to which a fuzzy relation $X$ is a solution to a WLS and a degree to which $X$ is equal to some other fuzzy relation $X'$ both determine the degree to which $X'$ is a solution to the WLS.

\begin{corollary}
Given a linearly ordered Heyting algebra $\cal L$ and $X, X' \in \fr{U}$, then for every $k \in \{1, 2, 3\}$ we have:
\begin{enumerate}
    \item[(a)] If $[SD_k(\fn{S})](X) < (X \gequiv X')$, then $[SD_k(\fn{S})](X') = [SD_k(\fn{S},X_0)](X)$;
    \item[(b)] Otherwise, if $[SD_k(\fn{S})](X) \ge (X \gequiv X')$, then $[SD_k(\fn{S})](X') \ge (X \gequiv X')$.
\end{enumerate}
\end{corollary}

Next, consider two WLSs given by families $\fn{S} = \{ R_i\}_{i \in I}$ and $\fn{S}' = \{ R'_i\}_{i \in I}$ defined over the universe $U$. Then the degree to which these two families are equal is:
\begin{equation*}
    (\fn{S} \gequiv \fn{S}') = \bigwedge_{i \in I} ( R_i \gequiv  R'_i).
\end{equation*}
We are now in a position to prove the following. For two WLSs that are equal to some degree, if a fuzzy relation solves one WLS to some degree, then it also solves the other WLS in a degree at least the conjunction of the previous two values.

\begin{lemma} \label{lem.SecLemaApproxSol}
Assume that $\cal L$ is a Heyting algebra and $X \in \fr{U}$. Moreover, fix $\fn{S} = \{ R_i\}_{i \in I}$ and $\fn{S}' = \{ R'_i\}_{i \in I}$ to be two families of fuzzy relations defined over $U$. Then for every $k \in \{1, 2, 3\}$ we have:
\begin{equation}
    (\fn{S} \gequiv \fn{S}') \land [SD_k(\fn{S})](X) = (\fn{S} \gequiv \fn{S}') \land [SD_k(\fn{S}')](X).
\end{equation}
\end{lemma}
\begin{proof}
We first focus to prove that:
\begin{equation}\label{eq.SecLemaApproxSol}
    (\fn{S} \gequiv \fn{S}') \land [SD_k(\fn{S},X_0)](X) \le [SD_k(\fn{S}',X_0)](X).
\end{equation}
In the same manner as we have done in the proof of Lemma~\ref{lem.FirstLemaApproxSol}, we start from the left-hand side of \eqref{eq.SecLemaApproxSol}, and use \eqref{eq.gsubgeq1}, \eqref{eq.gsubgeq2}, \eqref{eq.gsubgeq6} and \eqref{eq.gsubgeq8} to obtain
\begin{align*}
    (R \gequiv R') \land [SD_k(\fn{S},X_0)](X) 
    & = \left( \bigwedge_{i \in I} ( R_i \gequiv  R'_i) \right) \land 
    \left( \bigwedge_{i \in I} (X \circ  R_i \gsub  R_i \circ X) \right)\\
    & \le \bigwedge_{i \in I} [( R_i \gequiv  R'_i) \land (X \circ  R_i \gsub  R_i \circ X)] \\
    & = \bigwedge_{i \in I} [(X \gequiv X) \land ( R_i \gequiv  R'_i) \land ( R_i \gequiv  R'_i) \land (X \gequiv X) \land (X \circ  R_i \gsub  R_i \circ X)] \\
    & \le \bigwedge_{i \in I} [(X \circ  R_i \gequiv X \circ  R'_i) \land ( R_i \circ X \gequiv  R'_i \circ X) \land (X \circ  R_i \gsub  R_i \circ X)]\\
    & \le \bigwedge_{i \in I} (X \circ  R'_i \gsub  R'_i \circ X) \\
    & = [SD_k(\fn{S}',X_0)](X).
\end{align*}
Now the statement of the Lemma follows easily from \eqref{eq.SecLemaApproxSol}.
\end{proof}

Again, in a linearly ordered Heyting algebra, we conclude that the degree to which a fuzzy relation $X$ solves some WLS and a degree to which this WLS is equal to some other WLS both determine the degree to which $X'$ solves the other WLS.

\begin{corollary}
If $\cal L$ is a linearly ordered Heyting algebra, $X \in \fr{U}$, $\fn{S} = \{ R_i\}_{i \in I}$ and $\fn{S}' = \{ R'_i\}_{i \in I}$ are two families of fuzzy relations defined over $U$, then for every $k \in \{1, 2, 3\}$ we have:
\begin{enumerate}
    \item[(a)] If $[SD_k(\fn{S})](X) < (\fn{S} \gequiv \fn{S}')$, then $[SD_k(\fn{S}')](X) = [SD_k(\fn{S})](X)$;
    \item[(b)] Otherwise, if $[SD_k(\fn{S})](X) \ge (\fn{S} \gequiv \fn{S}')$, then $[SD_k(\fn{S}')](X) \ge (\fn{S} \gequiv \fn{S}')$.
\end{enumerate}
\end{corollary}

As a direct consequence of Lemmas \ref{lem.FirstLemaApproxSol} and \ref{lem.SecLemaApproxSol}, we conclude the following: if two WLSs are equal to some degree, a fuzzy relation is a solution to the one WLS \modiff{to} some degree, and the other fuzzy relation is equal to the first fuzzy relation \modiff{to} some degree, then the second fuzzy relation is a solution to the second WLS \modiff{to} a degree at least the conjunction of the previous three degrees. 

\begin{theorem}
Given a Heyting algebra $\cal L$, $X, X' \in \fr{U}$, and two families $\fn{S} = \{ R_i\}_{i \in I}$ and $\fn{S}' = \{ R'_i\}_{i \in I}$ of fuzzy relations  defined over $U$, for every $k \in \{1, 2, 3\}$ we have:
\begin{align*}
    (X \gequiv X') \land (\fn{S} \gequiv \fn{S}') \land [SD_k(\fn{S},X_0)](X) 
    = (X \gequiv X') \land (\fn{S} \gequiv \fn{S}') \land [SD_k(\fn{S}',X_0)](X').
\end{align*}
\end{theorem}

\begin{corollary}
Given a linearly ordered Heyting algebra $\cal L$, $X, X' \in \fr{U}$, and two families $\fn{S} = \{ R_i\}_{i \in I}$ and $\fn{S}' = \{ R'_i\}_{i \in I}$ of fuzzy relations  defined over $U$, for every $k \in \{1, 2, 3\}$ we have:
\begin{enumerate}
    \item[(a)] If $[SD_k(\fn{S})](X) < (X \gequiv X') \land (\fn{S} \gequiv \fn{S}')$, then $[SD_k(\fn{S}')](X') = [SD_k(\fn{S})](X)$;
    \item[(b)] Otherwise, if $[SD_k(\fn{S})](X) \ge (X \gequiv X') \land (\fn{S} \gequiv \fn{S}')$, then $[SD_k(\fn{S}')](X') \ge (X \gequiv X') \land (\fn{S} \gequiv \fn{S}')$.
\end{enumerate}
\end{corollary}

\section{Computation of the greatest approximate solutions to WLSs}
\label{sec.CGAS}

Theorem~\ref{thm.GreatestLambdaRegRel} states that there exists the greatest fuzzy relation which belongs to $\acut{x}{SD_k(\fn{S}, X_0)}$ and which is limited by a given fuzzy relation $X_0 \in \fr{U}$, for every $k \in \{1, 2, 3\}$ and $x \in L$. In this section, we develop an iterative procedure to compute such a fuzzy relation. We provide the algorithm only for the case $k=3$, since situations when $k=1$ or $k=2$ are also included in this case.

\begin{theorem}\label{th:compregrel}
Let $x \in L$. Consider the following procedure for computing the array $\{X_n\}_{n\in \mathbb{N}_0}$ of fuzzy relations on $U$:
\begin{align}
        &X_{n+1}=X_n\land \bigwedge_{i\in I} (x \to R_i\circ X_n)\lr R_i\land \bigwedge_{i\in I} R_i\rr(x \to X_n\circ  R_i),\label{eq:seq.affb.k}
\end{align}
for every $n\in \mathbb{N}_0$. The following holds:
\begin{enumerate}[(a)]
    \item $\{X_n\}_{n \in \mathbb{N}}$ is non-increasing.
    \item $X_k$ is the greatest element of $\acut{x}{SD_3(\fn{S}, X_0)}$ if and only if $X_k = X_{k+1}$.
    \item The procedure \eqref{eq:seq.affb.k} terminates when ${\cal L}(\fn{S}, x)$ is a finite subalgebra of $\cal L$.
\end{enumerate}
\end{theorem}
\begin{proof} 
\begin{enumerate}[(a)]
\item Easy to prove.
\item Let $k\in \mathbb{N}$ be an arbitrary number. If $X_k \in \acut{x}{SD_3(\fn{S}, X_0)}$, that means that the following holds:
\begin{equation*}
    x \le (X_k \circ R_i \gsub R_i \circ X_k) \ \textrm{ and } \ x \le (R_i \circ X_k \gsub X_k \circ R_i), \quad \textrm{ for every } i \in I,
\end{equation*}
which is equivalent to:
\begin{equation}\label{eq.tempEq0}
    x\otimes X_k \circ  R_i \le  R_i \circ X_k  \ \textrm{ and } \ 
    x\otimes  R_i \circ X_k \le X_k \circ  R_i, \quad \textrm{ for every } i \in I.
\end{equation}
According to the residuation properties for fuzzy relations \eqref{eq.rightAdjFuzzyRel} and \eqref{eq.leftAdjFuzzyRel}, this is further equivalent to:
\begin{equation} \label{eq.tempEq1}
X_k\leqslant \bigwedge_{i\in I} (x \to R_i\circ X_k)\lr R_i \ \textrm{ and } \  X_k\leqslant\bigwedge_{i\in I} R_i\rr(x \to X_k\circ  R_i),
\end{equation}
and since $X_k \le X_k$ always holds, we get that:
\begin{equation*}
    X_k \le X_k \land \bigwedge_{i\in I} (x \to R_i\circ X_k)\lr R_i \land \bigwedge_{i\in I} R_i\rr(x \to X_k\circ  R_i) = X_{k+1}.
\end{equation*}
Since  $X_{k+1}\le X_k$ follows from part (a), we have $X_{k+1}=X_k$.
Contrary, let $X_{k+1}=X_k$. ~Then \eqref{eq.tempEq1} follows, which means that:
\[
X_k\leqslant (x \to R_i\circ X_k) \lr R_i,\ \textrm{ and } \  X_k\leqslant R_i\rr(x \to X_k\circ  R_i), \quad \textrm{ for every } i \in I.
\]
Hence:
\[
X_k\circ R_i \leqslant x \to R_i\circ X_k,\ \textrm{ and } \  R_i\circ X\leqslant x \to X_k\circ  R_i, \quad \textrm{ for every } i \in I,
\] 
and according to the residuation properties for fuzzy relations \eqref{eq.rightAdjFuzzyRel} and \eqref{eq.leftAdjFuzzyRel}, these inequalities are equivalent \eqref{eq.tempEq0}, which further yields $X_k \in \acut{x}{SD_3(\fn{S}, X_0)}$.
        
To prove that $X_k$ is the greatest solution, we show that an arbitrary $Q \in \acut{x}{SD_3(\fn{S}, X_0))}$ is less than or equal to every fuzzy relation $X_n$ obtained by procedure \eqref{eq:seq.affb.k}. Indeed, by the definition of $\acut{x}{SD_3(\fn{S}, X_0)}$ we have $Q \le  X_0$. Suppose that $Q\le X_n $, for some $n\in\mathbb N_0$. Then for every $i\in I$, $x\otimes Q\circ  R_i\le  R_i\circ Q\le  R_i\circ X_n$, which implicate $Q\le (x \to R_i\circ X_n) \lr R_i$, and similarly $Q\le R_i \rr (x \to X_n\circ  R_i)$ for every $i\in I$, we have:
\[
Q \leqslant X_n\land \bigwedge_{i\in I}(x \to R_i\circ X_n) \lr R_i\wedge\bigwedge_{i\in I} R_i\rr (x \to X_n\circ  R_i)= X_{n+1}.
\]
According to the mathematical induction it follows $Q\leqslant X_n$, for every $n\in\mathbb N_0$, and hence $Q\leqslant X_k$.
Thus, $X_k$ is the greatest element of $\acut{x}{SD_3(\fn{S}, X_0)}$.

\item The number of different elements in the sequence $\{X_n\}_{n\in \mathbb{N}_0}$ must be finite since, by the assumption, ${\cal L}(\fn{S}, x)$ is a finite set. \qedhere
\end{enumerate}
\end{proof}

Theorem~\ref{th:compregrel} develops a method for computing the greatest element of $\acut{x}{SD_3(\fn{S}, X_0)}$, for a given $x \in L$. This method is formalized by Algorithm~\ref{alg1}. 

\begin{algorithm}[!htb]
\begin{algorithmic}[1]
\caption{Computation of the greatest element of $\acut{x}{SD_3(\fn{S}, X_0)}$}
\label{alg1}
\Require A family $\fn{S} = \{ R_i\}_{i \in I}$ of fuzzy relations, $X_0\in \fr{U}$, and $x \in L$;
\Ensure The greatest element of $\acut{x}{SD_3(\fn{S}, X_0)}$.
\State $X_1 \gets X_0$;
\Repeat
   \State $X_2 \gets X_1$;
    \State $X_1  \gets X_1 \land \bigwedge_{i\in I}(x \to R_i\circ  X_1) \lr R_i\wedge\bigwedge_{i\in I} R_i\rr(x \to X_1\circ  R_i)$;
\Until{$X_1 = X_2 $} 
\State \Return $X_2$.
\end{algorithmic}
\end{algorithm}

\modif{Let us determine the computation time of Algorithm~\ref{alg1}. We assume that the computation times for computing supremum, infimum, multiplication and residuum are all constants. In Step 4, we firstly compute compositions $R_i \circ X_1$ and $X_1 \circ R_i$, which can be done in $\bigo(|U|^3)$. After that, it takes $\bigo(|U|^2)$ time to compute $x \to R_i \circ X_1$ and $x \to X_1 \circ R_i$. In the end, the computation of residuals $(x \to R_i\circ  X_1) \lr R_i$ and $R_i\rr(x \to X_1\circ  R_i)$ also takes $\bigo(|U|^3)$ time. We take the infimum of all such fuzzy relations for every $i \in I$, which means that the computation time of Step 4 is $\bigo(|I||U|^3)$.

Since Steps 1 and 3 execute in constant time, the only thing remaining is to determine the number of times we iterate through the loop given by Steps 2 to 5. In each step of the loop, we memorize the fuzzy relation from the previous step in $X_2$, and then calculate a new fuzzy relation $X_1$ in $\bigo(|I||U|^3)$ time. After that, we check whether these two fuzzy relations $X_1$ and $X_2$ are equal, and if that is the case, we terminate the algorithm and return that fuzzy relation $X_1 = X_2$. We check the condition $X_1 = X_2$ by comparing all the values from fuzzy relations. Thus the algorithm terminates when all the values from $X_1$ and $X_2$ are equal.

Since the underlying structure of truth values is a complete residuated lattice, which is a very general algebraic structure, it can happen that we never reach a condition $X_1 = X_2$, or in other words, the set of values that the algorithm generates to populate the entries for the fuzzy relation $X_1$ can be infinite. The advantage of using a threshold $x$ in order to obtain an approximate solution is that we disregard the values greater than $x$ (because $x \to y = 1$ if and only if $x \le y$, all the values greater than or equal to $x$ in the entries of fuzzy relations $R_i \circ X_1$ and $X_1 \circ R_i$ become $1$). Thus, it can happen that the subalgebra ${\cal L}(\fn{S}, x)$ is finite, whether ${\cal L}(\fn{S})$ is not. However, since the threshold $x$ is in no way connected to the size $|U|$ of the universe $U$, we cannot choose an appropriate $x$ so that we obtain a finite subalgebra ${\cal L}(\fn{S}, x)$. However, depending on the context of the concrete system of fuzzy relations, we can choose some acceptable threshold $x$ so that the subalgebra ${\cal L}(\fn{S}, x)$ is finite, and Algorithm~\ref{alg1} terminates in a finite number of steps.

In the case when ${\cal L}(\fn{S}, x)$ is a finite subalgebra, denote with $l$ the number of elements of this set. Since the array $\{X_n\}_{n \in \mathbb{N}_0}$ is descending, every entry can change its value at most $l-1$ times. There are $|U|^2$ entries, which means that the complexity of the loop is $\bigo(l|U|^2)$. As we have calculated above, every loop step works in the $\bigo(|I||U|^3)$ complexity. That means that the total complexity of the algorithm is polynomial $\bigo(l|I||U|^5)$.}

The following example illustrates the situation when Algorithm~\ref{alg1} does not terminate when $x=1$, but it stops after three steps in the case when we seek a solution to the degree $x = 0.8$.

\begin{example}\label{exm.CompApproxSol}
Assume that a known fuzzy relation $R \in \fr{U}$ over the product structure in the weakly linear system \eqref{eq.WLS3} is equal to:
\[
R = 
\left[\begin{array}{cccccc}0.9&0&0&0&0.5&0\\0&0.8&0&0.3&0&0.2\\0&0&0.8&0.4&0&0.4\\0&0&0.8&0.2&0.2&0\\0&1&0&1&0.2&0\\0&0&0.9&0&0&0.1\\\end{array}\right].
\]
Set $X_0 = \blacktriangle_U$. In the case when we seek the exact greatest solution to \eqref{eq.WLS3} (that is, the case when $x=1$), Algorithm \ref{alg1} does not terminate. However, if we seek the greatest approximate solution in the degree $x=0.8$, the algorithm outputs the following fuzzy relations:
\begin{align*}
    X_1 = \left[\begin{array}{cccccc}1&1&1&1&25/36&5/9\\1&1&1&1&5/8&1/2\\1&1&1&1&25/36&5/9\\1&1&1&1&5/8&1/2\\1&1&1&1&1&1\\1&1&1&1&1&1\\\end{array}\right], \
    X_2 = X_3 = \left[\begin{array}{cccccc}1&1&1&625/648&25/36&5/9\\1&1&1&125/128&5/8&1/2\\1&1&1&125/128&25/36&5/9\\1&1&1&1&5/8&1/2\\1&1&1&1&1&1\\1&1&1&1&1&1\\\end{array}\right].
\end{align*}
According to the previous algorithm, we have that $X_2 = X_3$ is the greatest solution to the degree $x=0.8$ contained in $X_0 = \blacktriangle_U$ to \eqref{eq.WLS3}.
\end{example}

\section{Computation of approximate solutions to WLSs \modiff{that} are fuzzy preorders and fuzzy equivalences}
\label{sec.CGFP}

In Section \ref{sec.SD}, we have shown that there exists the greatest fuzzy relation that is a solution to the certain WLS in a chosen degree $x \in L$. 
Furthermore, we have provided the algorithm for its computation. 
However, in applications, we are usually not interested in arbitrary approximate solutions but in those approximate solutions that are fuzzy preorders or fuzzy equivalences.  
Recall that there always exists the greatest fuzzy preorder (fuzzy equivalence) that solves any WLS and is contained in a given fuzzy preorder (fuzzy equivalence) (cf. \cite{ICB.10}). Contrary to that, there may not exist the greatest fuzzy preorder (or the greatest fuzzy equivalence) that solves the observed WLS to a chosen degree contained in some fuzzy preorder (or fuzzy equivalence). We prove this fact in the following result.

\begin{proposition}\label{prop1}
It is not true that for every $k\in\{1,2,3\}$, $x\in X$ and for every fuzzy preorder $X_0\in \fp{U}$, there exists the greatest fuzzy preorder in the set $\acut{x}{SD_3(\fn{S}, X_0)}$. 
\end{proposition}
\begin{proof}
Our intention is to show that there exists $x\in X$ and a fuzzy preorder $X_0\in \fp{U}$, such that the set $\acut{x}{SD_3(\fn{S}, X_0)}$ has no greatest fuzzy preorder, which we do by an example.
Let $\fn{S}=\{ R_1\}$, where $ R_1$ is a fuzzy relation defined over the product structure as:
\[ R_1=\left[\begin{array}{ccc}0&0.4&1\\0.6&0.6&0.8\\0.8&1&1\\\end{array}\right].
\]
Moreover, let
\[
X_0=
\left[\begin{array}{ccc}1&0.2&0.06\\0.4&1&0.3\\0.24&0.6&1\\\end{array}\right].
\]
be a fuzzy preorder. Then for fuzzy preorders:
\[
X_1 =
\left[\begin{array}{ccc}1&0.2&0\\0.4&1&0\\0&0&1\\\end{array}\right],
\quad
X_2 =
\left[\begin{array}{ccc}1&0&0\\0&1&0.3\\0&0.6&1\\\end{array}\right],
\]we have that $X_1\in \acut{3/4}{SD_3(\fn{S}, X_0)}$ and $X_2\in \acut{2/3}{SD_3(\fn{S}, X_0)}$. Therefore, $X_1\vee X_2\in \acut{2/3}{SD_3(\fn{S}, X_0)}.$ 
It is straightforward to verify $X_0=(X_1\vee X_2)^{\infty},$ but $X_0\not\in\acut{2/3}{SD_3(\fn{S}, X_0)}$, because $[SD_3(\fn{S}, X_0)](X_0)=1/2.$
If we assume that $\acut{2/3}{SD_3(\fn{S}, X_0)}$ contains the greatest fuzzy preorder, denoted by $\tilde{X}$, then $X_1\le \tilde{X}$ and $X_2\le \tilde{X}$. This implies $X_1\vee X_2\le \tilde{X}$, and consequently $(X_1\vee X_2)^{\infty}\le \tilde{X}$, that is, $X_0\le \tilde{X}.$
Since $\tilde{X}$ is an element of $\acut{2/3}{SD_3(\fn{S}, X_0)}$, we have that $\tilde{X}\le X_0.$ Therefore, $\tilde{X}=X_0,$ which contradicts the fact $X_0\not\in\acut{2/3}{SD_3(\fn{S}, X_0)}$. This means that the set $\acut{2/3}{SD_3(\fn{S}, X_0)}$ has no greatest fuzzy preorder.
\end{proof}

As we show in this section, we can adapt Algorithm~\ref{alg1}, given in the previous section, to obtain an algorithm that calculates some fuzzy preorder that is an approximate solution to the WLS. Analogously, we can construct an analogous algorithm to calculate such fuzzy equivalence. 

The idea behind this adaptation lies in the following observations.
Consider the previous procedure in the case when $x = 1$, that is, when we compute the greatest exact solution to \eqref{eq.WLS3}. According to this procedure, we construct a sequence $\{X_n\}_{n \in \mathbb{N}_0}$ of fuzzy relations, and when the two successive fuzzy relations are equal, then we stop the procedure and say that such fuzzy relation is the greatest solution to \eqref{eq.WLS3}. And if $X_0$ is a fuzzy preorder, then the algorithm outputs the greatest fuzzy preorder that is a solution to \eqref{eq.WLS3}. The question is — can we relax the stopping criterion of the procedure so that we do not stop when two successive fuzzy relations are equivalent, but when they are equivalent to some desirable degree $x$? The answer is the following: yes, we can stop the procedure earlier when two successive fuzzy relations from the generated sequence are equal to some chosen degree $x \in L$, and the resulting fuzzy preorder is a solution to at least that degree $x$ to \eqref{eq.WLS3}. However, the drawback of this approach is that this fuzzy relation does not necessarily have to be a maximal such fuzzy relation. An additional modification must be made, so the algorithm outputs a fuzzy preorder in each step. We focus on developing this algorithm through the rest of this section. 

For every $ R \in \fr{U}$, define three fuzzy subsets $SD_4( R), SD_5( R)$ and $SD_6( R)$ on the set $\fr{U}$ as:
\begin{align}
    & [SD_4( R)](X) = (X \circ  R \circ X \gequiv  R \circ X), \\
    & [SD_5( R)](X) = (X \circ  R \circ X \gequiv X \circ  R), \\
    & [SD_6( R)](X) = [SD_4( R)](X) \land [SD_5( R)](X),
\end{align}
for every $X \in \fr{U}$. In addition, we define three additional fuzzy subsets $SD_7( R), SD_8( R)$ and $SD_9( R)$ on the set $\fr{U}$ as:
\begin{align}
    & [SD_7( R)](X) = (X \gsub ( R \circ X) \lr ( R \circ X)), \\
    & [SD_8( R)](X) = (X \gsub (X \circ  R) \rr (X \circ  R)), \\
    & [SD_9( R)](X) = [SD_7( R)](X) \land [SD_8( R)](X).
\end{align}
By the same token, for a given family $\fn{S} = \{ R_i\}_{i \in I}$, we define fuzzy subsets $SD_k(R)$, for every $k \in \{4, 5, 6, 7, 8, 9\}$, on the set $\fr{U}$ as 
\[
[SD_k(\fn{S})](X) = \bigwedge_{i \in I} [SD_k( R_i)](X),
\]
for every $X \in \fr{U}$. According to \eqref{eq.gsubgeq9}, we have that:
\begin{align}
    & [SD_7(\fn{S})](X) = \left(X \gsub \bigwedge_{i \in I}( R_i \circ X) \lr ( R_i \circ X)\right), \\
    & [SD_8(\fn{S})](X) = \left(X \gsub \bigwedge_{i \in I}(X \circ  R_i) \rr (X \circ  R_i)\right), \\
    & [SD_9(\fn{S})](X) = [SD_7(\fn{S})](X) \land [SD_8(\fn{S})](X).
\end{align}
Again, let $\acut{x}{SD_k(\fn{S}, X_0)} = \acut{x}{SD_k(\fn{S})} \cap \{ X \in \fr{U} | X \le X_0\}$, for every $k \in \{4,5,6,7,8,9\}$. To prove the equivalence of the corresponding sets $\acut{x}{SD_k(\fn{S}, X_0)}$, we firstly prove the following auxiliary result, which is a generalization of the residuation properties \eqref{eq.rightAdjFuzzyRel} and \eqref{eq.leftAdjFuzzyRel} for fuzzy relations.

\begin{lemma} \label{lem.AjdPropFuzzRelDegree}
For fuzzy relations $X, R, Q \in \fr{U}$, the following \emph{generalized residuation properties} hold:
\begin{align}
    & (X \circ R \gsub Q) = (X \gsub Q \lr R), \label{eq.LRDegree}\\
    & (R \circ X \gsub Q) = (X \gsub R \rr Q). \label{eq.RRDegree}
\end{align}
\end{lemma}
\begin{proof}
We prove only \eqref{eq.LRDegree}, since \eqref{eq.RRDegree} follows analogously. Starting from the left-hand side of \eqref{eq.LRDegree} we obtain
\begin{align*}
    (X \circ R \gsub Q) & = \bigwedge_{u \in U} \bigwedge_{v \in U} (X \circ R)(u, v) \to Q(u, v)  
    = \bigwedge_{u \in U} \bigwedge_{v \in U} \Bigg(\bigvee_{w \in U}X(u, w) \otimes R(w, v)\Bigg) \to Q(u, v).
\end{align*}
Properties \eqref{eq.CRLProp2} and \eqref{eq.CRLProp1} further yield
\begin{align*}
    (X \circ R \gsub Q) & = \bigwedge_{u \in U} \bigwedge_{v \in U} \bigwedge_{w \in U}[(X(u, w) \otimes R(w, v)) \to Q(u, v)] \\
    & = \bigwedge_{u \in U} \bigwedge_{v \in U} \bigwedge_{w \in U}[X(u, w) \to (R(w, v) \to Q(u, v))].
\end{align*}
In the end, \eqref{eq.CRLProp3} gives
\begin{align*}
    (X \circ R \gsub Q) & = \bigwedge_{u \in U} \bigwedge_{w \in U} X(u, w) \to \Bigg(\bigwedge_{v \in U}R(w, v) \to Q(u, v)\Bigg) \\
    & = \bigwedge_{u \in U} \bigwedge_{w \in U} X(u, w) \to (Q \lr R)(u, w) \\
    & = (X \gsub Q \lr R),
\end{align*}
which completes the proof.
\end{proof}

\begin{theorem} \label{thm.SystemsEquiv}
Let $P, X_0 \in \fp{U}$ be fuzzy preorders on $U$. Then, for every $x \in L$ and $k \in \{1, 2, 3\}$, the following statements are equivalent:
\begin{enumerate}[(a)]
    \item $P \in \acut{x}{SD_k(\fn{S}, X_0)}$,
    \item $P \in \acut{x}{SD_{k + 3}(\fn{S}, X_0)}$,
    \item $P \in \acut{x}{SD_{k + 6}(\fn{S}, X_0)}$.
\end{enumerate}
\end{theorem}
\begin{proof}
Recall that, if $P$ is a fuzzy preorder on $U$, then we have $P \circ P = P$.
We prove all the statements only in the case when $k = 1$, since in other cases the proof is similar. In all cases, we employ properties \eqref{eq.gsubgeq1}, \eqref{eq.gsubgeq4}, \eqref{eq.gsubgeq5} and \eqref{eq.gsubgeq7}.

First, we prove the first “if and only if'' statement. If $P \in \acut{x}{SD_1(\fn{S}, X_0)}$, then $x \le [SD_1(\fn{S}, X_0)](P)$. Further we have:
\begin{align*}
x &\le (P \circ  R_i \gsub  R_i \circ P) = (P \circ  R_i \gsub  R_i \circ P) \otimes (P \gsub P) 
\le (P \circ  R_i \circ P \gsub  R_i \circ P), \quad \textrm{ for every } i \in I.
\end{align*}
In addition we have that $ R_i \circ P \le P \circ  R_i \circ P$, for every $i \in I$, which means that $x \le 1 = ( R_i \circ P \gsub P \circ  R_i \circ P)$. Thus, it follows that $x \le [SD_4(\fn{S}, X_0)](P)$. Conversely, assume that $P \in \acut{x}{SD_4(\fn{S}, X_0)}$. Then from the fact that $P \circ  R_i \le P \circ  R_i \circ P$, for every $i \in I$, we have $(P \circ  R_i \gsub P \circ  R_i \circ P) = 1$. This means that for every $i \in I$ we have
\begin{align*}
x &\le (P \circ  R_i \circ P \gequiv  R_i \circ P) \\
& \le (P \circ  R_i \circ P \gsub  R_i \circ P) \\
& = (P \circ  R_i \gsub P \circ  R_i \circ P) \otimes (P \circ  R_i \circ P \gsub  R_i \circ P) \\
&\le (P \circ  R_i \gsub  R_i \circ P),
\end{align*}
which means that $P \in \acut{x}{SD_1(\fn{S}, X_0)}$. 

Now, we prove the second “if and only if'' statement. Assume that $P \in \acut{x}{SD_4(\fn{S}, X_0)}$. Then from Lemma~\ref{lem.AjdPropFuzzRelDegree} we conclude that the following holds:
\begin{align*}
    x & \le (P \circ  R_i \circ P \gequiv  R_i \circ P)
      \le (P \circ  R_i \circ P \gsub  R_i \circ P) 
      = (P \gsub ( R_i \circ P) \lr ( R_i \circ P)), \quad \textrm{ for every } i \in I.
\end{align*}
From this and \eqref{eq.gsubgeq9} it further follows that
\begin{align*}
    x & \le \Bigg(\bigwedge_{i \in I} P \gsub ( R_i \circ P) \lr ( R_i \circ P) \Bigg)
    = \Bigg( P \gsub \bigwedge_{i \in I} ( R_i \circ P) \lr ( R_i \circ P) \Bigg) = [SD_7(\fn{S}, X_0)](P),
\end{align*}
which means that $P \in \acut{x}{SD_7(\fn{S}, X_0)}$. Conversely, if $P \in \acut{x}{SD_7(\fn{S}, X_0)}$, then
\begin{align*}
    x &\le \Bigg( P \gsub \bigwedge_{i \in I} ( R_i \circ P) \lr ( R_i \circ P) \Bigg)
    = \Bigg(\bigwedge_{i \in I} P \gsub ( R_i \circ P) \lr ( R_i \circ P) \Bigg).
\end{align*}
This means that for every $i \in I$ we have
\begin{equation*}
    x \le (P \gsub (R_i \circ P) \lr (R_i \circ P)) = (P \circ R_i \circ P \gsub R_i \circ P).
\end{equation*}
As noted before, $x \le ( R_i \circ P \gsub P \circ  R_i \circ P)$ holds. Thus, $x \le (P \circ  R_i \circ P \gequiv  R_i \circ P)$, for every $i \in I$, and $P \in \acut{x}{SD_4(\fn{S}, X_0)}$ follows.
\end{proof}

Now we are in a position to develop a method to compute a fuzzy preorder that is a solution to a WLS in a certain degree. For that reason, for a given family $\fn{S} = \{ R_i\}_{i \in I}$, define three functions $\mathcal{F}_1, \mathcal{F}_2$ and $\mathcal{F}_3$ on the lattice of all fuzzy preorders on $U$ into itself by:
\begin{align}
    & \mathcal{F}_1(X) = \bigwedge_{i \in I} ( R_i \circ X) \lr ( R_i \circ X), \label{eq.F1Varphi}\\
    & \mathcal{F}_2(X) = \bigwedge_{i \in I} (X \circ  R_i) \rr (X \circ  R_i), \label{eq.F2Varphi}\\
    & \mathcal{F}_3(X) = \mathcal{F}_1(X) \land \mathcal{F}_2(X).\label{eq.F3Varphi}
\end{align}

\begin{theorem} \label{thm.FuzzyPreorderComp}
Let $X_0\in \fp{U}$ be a fuzzy preorder on $U$. For every $k \in \{1, 2, 3\}$, consider a sequence $\{X_n^k\}_{n\in \mathbb{N}_0}$ of fuzzy relations on $U$ defined by the following formula: 
\begin{align}
        &X_{n+1}^k = X_n^k \land \mathcal{F}_k(X_n^k),\label{procedPreord}
\end{align}
for every $n\in \mathbb{N}_0$. Then the following properties hold for every $k \in \{1, 2, 3\}$ and $n\in \mathbb{N}_0$:
\begin{enumerate}[(a)]
    \item $X_{n+1}^k \le X_n^k$;
    \item $X_n^k$ is a fuzzy preorder;
    \item For every $x \in L$, $X_n^k \in \acut{x}{SD_k(\fn{S}, X_0)}$ if and only if $x \le (X_n^k \gequiv X_{n+1}^k)$;
    \item If ${\cal L}(\{ R_i\}_{i \in I}, x)$ is locally finite subalgebra of $\cal L$, then there exists $n \in \mathbb{N}_0$ such that $x \le (X_n^k \gequiv X_{n+1}^k)$, for every $x \in L$.
\end{enumerate}
\end{theorem}
\begin{proof}
\begin{enumerate}[(a)]
\item Follows straightforward from the construction of the sequence \eqref{procedPreord}.
\item Follows from Lemmas~\ref{lem.AuxFuzzyRel1} and~\ref{lem.AuxFuzzyRel2}.
\item We prove the case when $k = 3$, because it implies other cases. Choose some $x \in L$. 

Assume that $X_n^k \in \acut{x}{SD_3(\fn{S}, X_0)}$. According to Theorem~\ref{thm.SystemsEquiv}, this is equivalent to $X_n^k \in \acut{x}{SD_9(\fn{S}, X_0)}$, which means that
\begin{align} 
    &x \le \left(X_n^k \gsub \bigwedge_{i \in I}( R_i \circ X_n^k) \lr ( R_i \circ X_n^k)\right), \label{eq.varphiNKXGraded1} \\
    &x \le \left(X_n^k \gsub \bigwedge_{i \in I}(X_n^k \circ  R_i) \rr (X_n^k \circ  R_i)\right). \label{eq.varphiNKXGraded2}
\end{align}
Since we also have $x \le 1 = (X_n^k \gsub X_n^k)$, from \eqref{eq.gsubgeq9} we conclude
\begin{align*}
    x & \le \left(X_n^k \gsub X_n^k \land \bigwedge_{i \in I}( R_i \circ X_n^k) \lr ( R_i \circ X_n^k) 
    \land \bigwedge_{i \in I}(X_n^k \circ  R_i) \rr (X_n^k \circ  R_i) \right) 
    = (X_n^k \gsub X_{n + 1}^k).
\end{align*}
In addition, from part (a) we conclude that $x \le 1 = (X_{n + 1}^k \gsub X_n^k)$. That means that $x \le (X_{n}^k \gequiv X_{n+1}^k)$, which was to be proved. Conversely, assume that $x \le (X_{n}^k \gequiv X_{n+1}^k)$. From this assumption, it follows that $x \le (X_{n}^k \gsub X_{n+1}^k)$, and from the construction of the sequence \eqref{procedPreord}, we conclude that \eqref{eq.varphiNKXGraded1} and \eqref{eq.varphiNKXGraded2} hold, or in other words, $X_n^k \in \acut{x}{SD_9(\fn{S}, X_0)}$. Again, by Theorem~\ref{thm.SystemsEquiv}, this is equivalent to $X_n^k \in \acut{x}{SD_3(\fn{S}, X_0)}$, which was to be proven.
\item From the fact that ${\cal L}(\{ R_i\}_{i \in I}, x)$ is locally finite, it follows that  the number of different elements in the sequence $\{X_n^k\}_{n\in \mathbb{N}_0}$ is finite. Hence, there exists $m\in  \mathbb{N}_0$, such that $x \le 1 = (X_m^k \gequiv X_{m+1}^k)$. \qedhere
\end{enumerate}
\end{proof}

The procedure for computing the element of $\acut{x}{SD_3(\fn{S}, X_0)}$ that is a fuzzy preorder, for a given $x \in L$, is formalized by Algorithm~\ref{alg2}. \modif{It can be easily verified that its computation time is the same as for Algorithm~\ref{alg1}.}

\begin{algorithm}[!htb]
\begin{algorithmic}[1]
\caption{Computation of the element of $\acut{x}{SD_3(\fn{S}, X_0)}$ which is a fuzzy preorder}
\label{alg2}
\Require A family $\fn{S} = \{ R_i\}_{i \in I}$ of fuzzy relations over $U$, a fuzzy preorder $X_0\in \fr{U}$, and $x \in L$;
\Ensure The element of $\acut{x}{SD_3(\fn{S}, X_0)}$ which is a fuzzy preorder.
\State $X_1 \gets X_0$;
\Repeat
   \State $X_2 \gets X_1$;
    \State $X_1  \gets X_1 \land \bigwedge_{i\in I}( R_i \circ X_1) \lr ( R_i \circ X_1) \wedge\bigwedge_{i\in I}(X_1 \circ  R_i) \rr (X_1 \circ  R_i)$;
\Until{$x \le (X_1 \gequiv X_2)$} 
\State \Return $X_2$.
\end{algorithmic}
\end{algorithm}

Consider the problem of computing a fuzzy equivalence belonging to the set $\acut{x}{SD_3(\fn{S}, X_0)}$, where $X_0$ is a fuzzy equivalence. In order to do this, we redefine mappings $\mathcal{F}_1, \mathcal{F}_2$ and $\mathcal{F}_3$ to be defined on the lattice of all fuzzy equivalences on $U$ into itself, by replacing $\rr$ with $ \rre$, as well as $\lr$ with $\lre$ in \eqref{eq.F1Varphi}--\eqref{eq.F3Varphi}. With this in mind, Theorem~\ref{thm.FuzzyPreorderComp} states in part (b) that $X_n^k$ is a fuzzy equivalence. 
Algorithm~\ref{alg2} can be adapted accordingly.

\section{Applications in aggregation of fuzzy networks}
\label{sec.App}

One can define a \emph{fuzzy network} in many ways, according to the concrete approach to studying them. From the aspect of graph theory, we visualize a fuzzy network as \modiff{a} set of nodes and represent the connections between nodes by the edges between them, where we assign one or multiple labels from some appropriate set of truth values to each edge. An equivalent definition comes from the algebraic aspect, where the nodes are a universe set, and we represent the connections between nodes by the family of fuzzy relations on the universe. In what follows,  
a \emph{fuzzy network} is an ordered pair $\fn{S} = (U, \{ R_i\}_{i \in I})$ such that $U$ is a finite nonempty set and $\{ R_i \}_{i \in I}$ is a family of fuzzy relations on $U$. Then the \emph{nodes} of the fuzzy network $\fn{S}$ are the elements of the set $U$, while the family $\{ R_i\}_{i\in I}$ represents the connections among the nodes. We can capture the vagueness in the connections between nodes by using fuzzy relations to model these connections.

Given a fuzzy equivalence on the set of nodes of a fuzzy network, we can group these nodes to obtain the aggregated version of this fuzzy network according to the equivalence classes of a given fuzzy equivalence in the following way. 
Let $\fn{S}=(U,\{ R_i\}_{i \in I})$ be a fuzzy network and $X$ a fuzzy preorder on $U$. Recall that the natural fuzzy equivalence of $X$ is denoted by $\nateq{X}$. Now we are in a position to define a fuzzy network 
$\fn{S}/\nateq{X}=(U/\nateq{X},\{\nateq{ R}_i\}_{i \in I})$. The nodes of this fuzzy network are the equivalence classes of $\nateq{X}$, and ${\nateq{R}}_i\in \fr{U/\nateq{X}}$ is a fuzzy relation on $U/\nateq{X}$ defined for each $i \in I$ by:
\begin{equation}\label{eq.factorfuzzyrelation1}
    {\nateq{ R}}_i(\nateq{X}_u,\nateq{X}_v)=(X\circ R_i\circ X)(u,v)
\end{equation}
for all $u,v\in U$. It is not hard to prove that equation \eqref{eq.factorfuzzyrelation1} is equivalent to:
\begin{equation*}
    {\nateq{ R}}_i(\nateq{X}_u,\nateq{X}_v)=(u X)\circ R_i\circ(X v)
\end{equation*}
where $u X$ is $X$-afterset of $u$ and $X v$ is $X$-foreset of $v$. It is not hard to check that ${\nateq{R}}_i$ is a well-defined mapping. We can analogously define fuzzy networks $X \fn{S}$ and $\fn{S} X$, where we replace the set $U/{\nateq{X}}$ by the set of aftersets or foresets of $X$. In such fuzzy networks nothing changes since the definition \eqref{eq.factorfuzzyrelation1} is preserved. It follows that $\fn{S}/\nateq{X}$, $X \fn{S}$ and $\fn{S} X$ are mutually isomorphic to each other (cf. \cite{SCI.17}). In what follows, we call the fuzzy network $\fn{S}/\nateq{X}$ the \emph{factor fuzzy network of $\fn{S}$ with respect to $X$}.

The ultimate goal in studying fuzzy networks is to find their aggregated alternatives by using factor fuzzy networks obtained by the previous definition. In other words, the aggregated version of a fuzzy network is another fuzzy network that is smaller than the original fuzzy network and preserves the connections between the nodes. To achieve this, we do not employ arbitrary fuzzy preorders and fuzzy equivalences to build the factor fuzzy network. Many such fuzzy relations preserve the connections between nodes, and one of them is the definition of the so-called \emph{regular fuzzy equivalences}. Recall that Fan et al. \cite{FL.14,FLL.07,FLL.08} defined \modiff{regular fuzzy} equivalences as fuzzy equivalences that solve the weakly linear system \eqref{eq.WLS3}. By this definition, two nodes are \emph{regularly equivalent} if they are connected to the same neighborhoods. With this in mind, we say that a fuzzy preorder (fuzzy equivalence) is a \emph{regular fuzzy preorder} (\emph{regular fuzzy equivalence}) if it is a solution to \eqref{eq.WLS3}.

\modif{The main problem when computing regular fuzzy equivalences for fuzzy networks is not that the methods for their computation may not terminate after a finite number of iterations. Instead, it is that} there are many cases when the only regular fuzzy equivalence that exists for a given fuzzy network $\fn{S} = (U, \{R_i\}_{i \in I})$ is the identity relation $\vartriangle_{U}$. Nevertheless, we can still group the nodes of a fuzzy network not to how they are regularly equivalent but to how they are regularly equivalent to some extent. In this way, just by relaxing the criteria for grouping the nodes of a fuzzy network, we can group the nodes and consequently construct a smaller aggregated fuzzy network. \modif{Note that Algorithm~\ref{alg2} outputs a fuzzy preorder which is, in general, neither maximal nor the greatest regular fuzzy preorder to some extent. Hence, the degree of equality of this fuzzy preorder to the greatest exact regular fuzzy preorder is of no interest for the aggregation of a fuzzy network. In fact, by choosing the extent to which we allow to break the connections between nodes of a fuzzy network and group them up to that extent, we can aggregate the fuzzy network so that it is satisfactorily small. The further we relax this degree of relaxation, the smaller the aggregated fuzzy network is in its size, but we must carefully choose this degree so that it is not too small because it will completely destroy the structure of the fuzzy network.}

\begin{example}
Let $X_0$ be a fuzzy relation and $\fn{S}=(U, R)$ a fuzzy network
given in Proposition~\ref{prop1}. Recall that the underlying structure is the product structure. Choose $\modif{x} = 3/4$. Then Algorithm~\ref{alg2} outputs the following fuzzy preorder:
\begin{equation*}
    X = \begin{bmatrix}  
        1 & 0.2 & 0.06\\
        0.3 & 1 & 0.3\\
        0.12 & 0.4 & 1
    \end{bmatrix}.
\end{equation*}
Note that $X$ is incomparable to both $X_1$ and $X_2$, so there is no greatest regular fuzzy preorder in the degree $3/4$ for this fuzzy network. 
\end{example}

In the following example, we depict a scenario when it is impossible to determine the greatest (exact) fuzzy preorder that is a solution to \eqref{eq.WLS3}, but Algorithm~\ref{alg2} finishes when computing a fuzzy preorder that is an approximate solution to \eqref{eq.WLS3}.

\begin{example}\label{exm.exmFN1}
Let $\fn{S}=(U, R)$ be a fuzzy network defined over the product structure given in Example~\ref{exm.CompApproxSol}.
Also, let $X_0 = \blacktriangle_U$. If we set $x=1$, then Algorithm~\ref{alg2} does not terminate. However, if we set $x=0.8$, then Algorithm~\ref{alg2} outputs fuzzy preorders given in Table~\ref{tbl.FP1}.

\begin{table}[htb]
\centering
\begin{tabular}{ccc}
  \toprule
     Step \# & $X_i$ & $X_{i-1} \gequiv X_{i}$\\
  \midrule
     $1$ & 
     $
     X_1 = \left[\begin{array}{cccccc}1&1&1&1&\frac{5}{9}&\frac{4}{9}\\\frac{8}{9}&1&\frac{9}{10}&1&\frac{1}{2}&\frac{2}{5}\\\frac{8}{9}&1&1&1&\frac{5}{9}&\frac{4}{9}\\\frac{8}{9}&1&\frac{9}{10}&1&\frac{1}{2}&\frac{2}{5}\\1&1&1&1&1&\frac{4}{5}\\1&1&1&1&\frac{9}{10}&1\\\end{array}\right]
     $ & $X_0 \gequiv X_1 = \frac{2}{5}$ \\
     $2$ & 
     $
     X_2=\left[\begin{array}{cccccc}1&\frac{8}{9}&\frac{8}{9}&\frac{50}{81}&\frac{5}{9}&\frac{4}{9}\\\frac{64}{81}&1&\frac{4}{5}&\frac{5}{8}&\frac{1}{2}&\frac{2}{5}\\\frac{64}{81}&1&1&\frac{5}{8}&\frac{5}{9}&\frac{4}{9}\\\frac{64}{81}&1&\frac{4}{5}&1&\frac{1}{2}&\frac{2}{5}\\\frac{80}{81}&1&1&1&1&\frac{4}{5}\\\frac{8}{9}&1&1&1&\frac{9}{10}&1\\\end{array}\right]
     $ & $X_1 \gequiv X_2 = \frac{50}{81}$ \\
     $3$ & 
     $
     X_3=\left[\begin{array}{cccccc}1&\frac{64}{81}&\frac{64}{81}&\frac{50}{81}&\frac{5}{9}&\frac{32}{81}\\\frac{512}{729}&1&\frac{4}{5}&\frac{5}{8}&\frac{40}{81}&\frac{2}{5}\\\frac{512}{729}&1&1&\frac{5}{8}&\frac{40}{81}&\frac{2}{5}\\\frac{512}{729}&1&\frac{4}{5}&1&\frac{40}{81}&\frac{2}{5}\\\frac{640}{729}&1&1&1&1&\frac{32}{45}\\\frac{64}{81}&1&1&1&\frac{9}{16}&1\\\end{array}\right]
     $ & $X_2 \gequiv X_3 = \frac{5}{8}$ \\
     $4$ & 
     $
     X_4=\left[\begin{array}{cccccc}1&\frac{512}{729}&\frac{512}{729}&\frac{50}{81}&\frac{5}{9}&\frac{256}{729}\\\frac{4096}{6561}&1&\frac{4}{5}&\frac{50}{81}&\frac{320}{729}&\frac{2}{5}\\\frac{4096}{6561}&\frac{8}{9}&1&\frac{50}{81}&\frac{320}{729}&\frac{2}{5}\\\frac{4096}{6561}&1&\frac{4}{5}&1&\frac{320}{729}&\frac{2}{5}\\\frac{5120}{6561}&1&1&1&1&\frac{256}{405}\\\frac{512}{729}&1&\frac{9}{10}&1&\frac{9}{16}&1\\\end{array}\right]
     $ & $X_3 \gequiv X_4 = \frac{8}{9}$ \\
     \bottomrule
\end{tabular}
\caption{Output of Algorithm~\ref{alg2} running on Example~\ref{exm.exmFN1}.}
\label{tbl.FP1}
\end{table}

Since we have $X_4 \gequiv X_3 = 8/9 > 0.8 = x$, Algorithm~\ref{alg2} stops after four steps and outputs the fuzzy relation $X_3$ as a fuzzy preorder that is a solution to \eqref{eq.WLS3} to the degree $x$.
\end{example}

In the following example, we show that there are situations where an aggregation of a fuzzy network can be achieved only by approximate solutions to \eqref{eq.WLS3}.

\begin{example}
Let $\fn{S}=(U, R)$ be a fuzzy network defined over the product structure given in Example~\ref{exm.CompApproxSol}.
Start from the universal relation $X_0 = \blacktriangle_U$.
If we choose the approximation degree $x=0.4$, then Algorithm~\ref{alg2} outputs $X_0$ as a resulting fuzzy preorder. In other words, we have chosen an approximation degree to be too low. If we increase the approximation degree to be $x=0.6$, then Algorithm~\ref{alg2} outputs the following fuzzy preorder:
\begin{equation}
\label{eq.FPX1}
X^{(1)} = \left[\begin{array}{cccccc}1&1&1&1&\frac{5}{9}&\frac{4}{9}\\\frac{8}{9}&1&\frac{9}{10}&1&\frac{1}{2}&\frac{2}{5}\\\frac{8}{9}&1&1&1&\frac{5}{9}&\frac{4}{9}\\\frac{8}{9}&1&\frac{9}{10}&1&\frac{1}{2}&\frac{2}{5}\\1&1&1&1&1&\frac{4}{5}\\1&1&1&1&\frac{9}{10}&1\\\end{array}\right].
\end{equation}
Since the second and the fourth row of $X^{(1)}$ are the same (which is equivalent to the fact that the second and the fourth column of $X^{(1)}$ are the same), we conclude that we can merge the second and the fourth node of the fuzzy network $\fn{S}$ to obtain the aggregated fuzzy network. We do not significantly reduce, but we still reduce a fuzzy network via an approximate solution. In the end, if we seek for the greatest fuzzy preorder that is an exact solution to \eqref{eq.WLS3} (case when the approximation degree has its greatest value $x=1$), then Algorithm~\ref{alg2} does not terminate. Suppose we employ any of the alternative ways to obtain this exact solution (for example, to take a limit of the convergent sequence generated by the Algorithm, see \cite{MJS.18} for more details). In that case, we obtain a fuzzy relation in which all rows (i.e., all columns) are mutually different. That means we cannot achieve a reduction of the fuzzy network $\fn{S}$ if we employ the fuzzy preorder that is an exact solution to \eqref{eq.WLS3}.
\end{example}

In the following example, we depict a situation where Algorithm~\ref{alg2} does not output a maximal fuzzy preorder. However, the obtained factor fuzzy network still gives a good reduction of the starting fuzzy network.

\begin{example}
Let $\fn{S}=(U, R)$ be a fuzzy network defined over the product structure given in Example~\ref{exm.CompApproxSol}.
Set $x = 0.6$ and $X_0 = \blacktriangle_U$. Then Algorithm~\ref{alg2} outputs the fuzzy preorder $X^{(1)}  \in \acut{x}{SD_3(\fn{S}, X_0)}$, given by \eqref{eq.FPX1}.
However, we also have that the fuzzy preorder $X^{(2)}$ is also an element of $\acut{x}{SD_3(\fn{S}, X_0)}$, where $X^{(2)}$ is given by:
\begin{equation*}
    X^{(2)} = \left[
        \begin{array}{cccccc}
            1 & 1 & 1 & 1 & 3/5 & 3/5 \\
            9/10 & 1 & 9/10 & 1 & 27/50 & 27/50 \\
            9/10 & 1 & 1 & 1 & 3/5 & 27/50 \\
            9/10 & 1 & 9/10 & 1 & 27/50 & 27/50 \\
            1 & 1 & 1 & 1 & 1 & 4/5 \\
            1 & 1 & 1 & 1 & 9/10 & 1 
        \end{array}
    \right].
\end{equation*}
It is evident that $X^{(2)} > X^{(1)}$, so Algorithm~\ref{alg2} doesn't necessarily produce a maximal or the greatest fuzzy preorder which is an element of the set $\acut{x}{SD_3(\fn{S}, X_0)}$. However, although Algorithm~\ref{alg2} does not produce a fuzzy relation greater than $X^{(2)}$, we see that factor networks $\fn{S} X^{(1)}$ and $\fn{S} X^{(2)}$ have the same number of states (because fuzzy relations $X^{(1)}$ and $X^{(2)}$ have the same number of different rows, or equivalently, the same number of different columns). Thus, we can indeed use Algorithm~\ref{alg2} to find a fuzzy preorder according to which we construct the factor fuzzy network.
\end{example}

\section{Conclusions}
This paper has studied approximate solutions to weakly linear systems of fuzzy relational equations and inequalities over complete residual lattices. Although such systems always have at least a trivial solution, it is a common situation that it is the only solution that exists. Therefore, we have described the set of all fuzzy relations that solve the system to a certain degree. In addition, for a predetermined degree of accuracy, we have given the algorithm for computing the greatest solution to that degree. We have also studied those approximate solutions that are fuzzy preorders and fuzzy equivalences. We have shown that, in the general case, the greatest such fuzzy relation does not have to exist. For this reason, we have proposed an algorithm that computes some approximate solution which is a fuzzy preorder. 

\modiff{The algorithms developed in this paper are iterative ones that utilize the Kleene Fixed Point Theorem. One possible direction in future research is to employ the well-known partition refinement technique to obtain faster algorithms when seeking solutions to weakly linear systems that are fuzzy equivalences or fuzzy preorders. Note that this technique has already given faster algorithms that compute crisp bisimulations for fuzzy automata (cf. \cite{MJS.18}). Moreover, s}ince the way to compute approximate solutions that are maximal fuzzy preorders is still an open problem, one direction in future work will be to develop a new methodology \modiff{to} solve this problem. Furthermore, another direction in our future work will be to study various generalizations of weakly linear systems studied in this research. \modif{Finally, although our iterative methods can finish in a finite number of iterations in cases when the previously developed ones do not, there does not exist a guarantee that they finish in a finite number of iterations for every underlying structure and every solution degree. A possible direction in future research can be to develop a new methodology to overcome this problem.}






\bibliographystyle{abbrvnat}
\bibliography{references.bib}

\begin{thebibliography}{68}
\providecommand{\natexlab}[1]{#1}
\providecommand{\url}[1]{\texttt{#1}}
\expandafter\ifx\csname urlstyle\endcsname\relax
  \providecommand{\doi}[1]{doi: #1}\else
  \providecommand{\doi}{doi: \begingroup \urlstyle{rm}\Url}\fi

\bibitem[{Abbasi Molai}(2010)]{A.10}
A.~{Abbasi Molai}.
\newblock Fuzzy linear objective function optimization with fuzzy-valued
  max-product fuzzy relation inequality constraints.
\newblock \emph{Mathematical and Computer Modelling}, 51\penalty0 (9):\penalty0
  1240--1250, 2010.

\bibitem[{Abbasi Molai}(2012)]{A.12}
A.~{Abbasi Molai}.
\newblock The quadratic programming problem with fuzzy relation inequality
  constraints.
\newblock \emph{Computers \& Industrial Engineering}, 62\penalty0 (1):\penalty0
  256--263, 2012.

\bibitem[{Abbasi Molai}(2014)]{A.14}
A.~{Abbasi Molai}.
\newblock A new algorithm for resolution of the quadratic programming problem
  with fuzzy relation inequality constraints.
\newblock \emph{Computers \& Industrial Engineering}, 72:\penalty0 306--314,
  2014.

\bibitem[b.~Qu et~al.(2015)b.~Qu, Sun, and f.~Wang]{QSW.15}
X.~b.~Qu, F.~Sun, and T.~f.~Wang.
\newblock Matrix elementary transformations in solving systems of fuzzy
  relation equations.
\newblock \emph{Applied Soft Computing}, 31:\penalty0 25--29, 2015.

\bibitem[Bartl(2015)]{B.15}
E.~Bartl.
\newblock Minimal solutions of generalized fuzzy relational equations:
  Probabilistic algorithm based on greedy approach.
\newblock \emph{Fuzzy Sets and Systems}, 260:\penalty0 25--42, 2015.

\bibitem[Bartl and Belohlavek(2011)]{BB.11}
E.~Bartl and R.~Belohlavek.
\newblock Sup-t-norm and inf-residuum are a single type of relational
  equations.
\newblock \emph{International Journal of General Systems}, 40\penalty0
  (6):\penalty0 599--609, 2011.

\bibitem[Bartl and Klir(2014)]{BK.14}
E.~Bartl and G.~J. Klir.
\newblock Fuzzy relational equations in general framework.
\newblock \emph{International Journal of General Systems}, 43\penalty0
  (1):\penalty0 1--18, 2014.

\bibitem[B{\v e}lohl{\' a}vek(2002)]{B.02a}
R.~B{\v e}lohl{\' a}vek.
\newblock \emph{Fuzzy Relational Systems: Foundations and Principles}.
\newblock Kluwer, New York, 2002.

\bibitem[B{\v e}lohl{\' a}vek and Vychodil(2005)]{BV.05}
R.~B{\v e}lohl{\' a}vek and V.~Vychodil.
\newblock \emph{Fuzzy Equational Logic, Studies in Fuzziness and Soft
  Computing}.
\newblock Springer, Berlin-Heidelberg, 2005.

\bibitem[Chung and Lee(1997)]{CL.97}
F.-L. Chung and T.~Lee.
\newblock A new look at solving a system of fuzzy relational equations.
\newblock \emph{Fuzzy Sets and Systems}, 88\penalty0 (3):\penalty0 343--353,
  1997.

\bibitem[{\'C}iri{\'c} and Bogdanovi{\'c}(2010)]{CB.10}
M.~{\'C}iri{\'c} and S.~Bogdanovi{\'c}.
\newblock Fuzzy social network analysis.
\newblock \emph{Godi{\v{s}}njak U{\v{c}}iteljskog fakulteta u Vranju},
  1:\penalty0 179--190, 2010.

\bibitem[{\' C}iri{\' c} et~al.(2007){\' C}iri{\' c}, Ignjatovi{\' c}, and
  Bogdanovi{\' c}]{CIB.07}
M.~{\' C}iri{\' c}, J.~Ignjatovi{\' c}, and S.~Bogdanovi{\' c}.
\newblock Fuzzy equivalence relations and their equivalence classes.
\newblock \emph{Fuzzy Sets and Systems}, 158\penalty0 (12):\penalty0 1295 --
  1313, 2007.

\bibitem[\'{C}iri\'{c} et~al.(2007)\'{C}iri\'{c}, Stamenkovi\'{c},
  Ignjatovi\'{c}, and Petkovi\'{c}]{CSIP.07}
M.~\'{C}iri\'{c}, A.~Stamenkovi\'{c}, J.~Ignjatovi\'{c}, and T.~Petkovi\'{c}.
\newblock Factorization of fuzzy automata.
\newblock In \emph{Proceedings of the 16th International Conference on
  Fundamentals of Computation Theory}, FCT'07, pages 213--225. Springer-Verlag,
  2007.

\bibitem[{\' C}iri{\' c} et~al.(2010){\' C}iri{\' c}, Stamenkovi{\' c},
  Ignjatovi{\' c}, and Petkovi{\' c}]{CSIP.10}
M.~{\' C}iri{\' c}, A.~Stamenkovi{\' c}, J.~Ignjatovi{\' c}, and T.~Petkovi{\'
  c}.
\newblock Fuzzy relation equations and reduction of fuzzy automata.
\newblock \emph{Journal of Computer and System Sciences}, 76\penalty0
  (7):\penalty0 609 -- 633, 2010.

\bibitem[{\' C}iri{\' c} et~al.(2012{\natexlab{a}}){\' C}iri{\' c},
  Ignjatovi{\' c}, Damljanovi{\' c}, and Ba{\v s}i{\' c}]{CIDB.12}
M.~{\' C}iri{\' c}, J.~Ignjatovi{\' c}, N.~Damljanovi{\' c}, and M.~Ba{\v
  s}i{\' c}.
\newblock Bisimulations for fuzzy automata.
\newblock \emph{Fuzzy Sets and Systems}, 186\penalty0 (1):\penalty0 100 -- 139,
  2012{\natexlab{a}}.

\bibitem[{\' C}iri{\' c} et~al.(2012{\natexlab{b}}){\' C}iri{\' c},
  Ignjatovi{\' c}, Jan{\v c}i{\' c}, and Damljanovi{\' c}]{CIJD.12}
M.~{\' C}iri{\' c}, J.~Ignjatovi{\' c}, I.~Jan{\v c}i{\' c}, and
  N.~Damljanovi{\' c}.
\newblock Computation of the greatest simulations and bisimulations between
  fuzzy automata.
\newblock \emph{Fuzzy Sets and Systems}, 208:\penalty0 22 -- 42,
  2012{\natexlab{b}}.

\bibitem[D{\' i}az-Moreno et~al.(2017)D{\' i}az-Moreno, Medina, and
  Turunen]{DMMT.17}
J.~C. D{\' i}az-Moreno, J.~Medina, and E.~Turunen.
\newblock Minimal solutions of general fuzzy relation equations on linear
  carriers. an algebraic characterization.
\newblock \emph{Fuzzy Sets and Systems}, 311\penalty0 (C):\penalty0 112--123,
  2017.

\bibitem[Esakia(2019)]{E.19}
L.~Esakia.
\newblock \emph{Heyting algebras: Duality theory}, volume~50.
\newblock Springer, 2019.

\bibitem[{Fan} et~al.(2007){Fan}, {Liau}, and {Lin}]{FLL.07}
T.~{Fan}, C.~{Liau}, and T.~{Lin}.
\newblock Positional analysis in fuzzy social networks.
\newblock In \emph{2007 IEEE International Conference on Granular Computing
  (GRC 2007)}, pages 423--423, 2007.

\bibitem[Fan and Liau(2014)]{FL.14}
T.-F. Fan and C.-J. Liau.
\newblock Logical characterizations of regular equivalence in weighted social
  networks.
\newblock \emph{Artificial Intelligence}, 214:\penalty0 66 -- 88, 2014.

\bibitem[Fan et~al.(2008)Fan, Liau, and Lin]{FLL.08}
T.-F. Fan, C.-J. Liau, and T.-Y. Lin.
\newblock A theoretical investigation of regular equivalences for fuzzy graphs.
\newblock \emph{International Journal of Approximate Reasoning}, 49\penalty0
  (3):\penalty0 678 -- 688, 2008.

\bibitem[Gottwald(1994)]{G.94}
S.~Gottwald.
\newblock Approximately solving fuzzy relation equations: Some mathematical
  results and some heuristic proposals.
\newblock \emph{Fuzzy Sets and Systems}, 66\penalty0 (2):\penalty0 175--193,
  1994.

\bibitem[Ignjatovi{\'c} and {\'C}iri{\'c}(2012)]{IC.12}
J.~Ignjatovi{\'c} and M.~{\'C}iri{\'c}.
\newblock Weakly linear systems of fuzzy relation inequalities and their
  applications: A brief survey.
\newblock \emph{Filomat}, 26\penalty0 (2):\penalty0 207--241, 2012.

\bibitem[Ignjatovi{\'c} et~al.(2010)Ignjatovi{\'c}, {\'C}iri{\'c}, and
  Bogdanovi{\'c}]{ICB.10}
J.~Ignjatovi{\'c}, M.~{\'C}iri{\'c}, and S.~Bogdanovi{\'c}.
\newblock On the greatest solutions to weakly linear systems of fuzzy relation
  inequalities and equations.
\newblock \emph{Fuzzy Sets and Systems}, 161\penalty0 (24):\penalty0
  3081--3113, 2010.

\bibitem[Ignjatovi{\'c} et~al.(2012)Ignjatovi{\'c}, {\'C}iri{\'c},
  Damljanovi{\'c}, and Jan{\v{c}}i{\'c}]{ICDJ.12}
J.~Ignjatovi{\'c}, M.~{\'C}iri{\'c}, N.~Damljanovi{\'c}, and
  I.~Jan{\v{c}}i{\'c}.
\newblock Weakly linear systems of fuzzy relation inequalities: The
  heterogeneous case.
\newblock \emph{Fuzzy Sets and Systems}, 199:\penalty0 64--91, 2012.

\bibitem[Ignjatovi{\' c} et~al.(2015)Ignjatovi{\' c}, {\' C}iri{\' c}, {\v
  S}e{\v s}elja, and Tepav{\v c}evi{\' c}]{ICST.15}
J.~Ignjatovi{\' c}, M.~{\' C}iri{\' c}, B.~{\v S}e{\v s}elja, and A.~Tepav{\v
  c}evi{\' c}.
\newblock Fuzzy relational inequalities and equations, fuzzy quasi-orders,
  closures and openings of fuzzy sets.
\newblock \emph{Fuzzy Sets and Systems}, 260:\penalty0 1--24, 2015.

\bibitem[Li and Yang(2012)]{LY.12}
J.-X. Li and S.~J. Yang.
\newblock Fuzzy relation inequalities about the data transmission mechanism in
  bittorrent-like peer-to-peer file sharing systems.
\newblock In \emph{2012 9th International Conference on Fuzzy Systems and
  Knowledge Discovery}, pages 452--456, 2012.

\bibitem[Li and p.~Wang(2021)]{LW.21}
M.~Li and X.~p.~Wang.
\newblock Remarks on minimal solutions of fuzzy relation inequalities with
  addition-min composition.
\newblock \emph{Fuzzy Sets and Systems}, 410:\penalty0 19--26, 2021.

\bibitem[Lin and Yang(2020)]{LY.20}
H.~Lin and X.~Yang.
\newblock Dichotomy algorithm for solving weighted min-max programming problem
  with addition-min fuzzy relation inequalities constraint.
\newblock \emph{Computers \& Industrial Engineering}, 146:\penalty0 106537,
  2020.

\bibitem[Luoh and Liaw(2010)]{LL.10}
L.~Luoh and Y.-K. Liaw.
\newblock Novel approximate solving algorithm for fuzzy relational equations.
\newblock \emph{Mathematical and Computer Modelling}, 52\penalty0 (1):\penalty0
  303--308, 2010.

\bibitem[Markovskii(2004)]{M.04}
A.~V. Markovskii.
\newblock Solution of fuzzy equations with max-product composition in inverse
  control and decision making problems.
\newblock \emph{Automation and Remote Control}, 65\penalty0 (9):\penalty0
  1486–1495, 2004.

\bibitem[Mici{\' c} et~al.(2018)Mici{\' c}, Jan{\v c}i{\' c}, and
  Stanimirovi{\' c}]{MJS.18}
I.~Mici{\' c}, Z.~Jan{\v c}i{\' c}, and S.~Stanimirovi{\' c}.
\newblock Computation of the greatest right and left invariant fuzzy
  quasi-orders and fuzzy equivalences.
\newblock \emph{Fuzzy Sets and Systems}, 339:\penalty0 99 -- 118, 2018.

\bibitem[p.~Wang(2001)]{W.01}
X.~p.~Wang.
\newblock Method of solution to fuzzy relation equations in a complete
  brouwerian lattice.
\newblock \emph{Fuzzy Sets and Systems}, 120\penalty0 (3):\penalty0 409--414,
  2001.

\bibitem[p.~Wang and Zhao(2013)]{WZ.13}
X.~p.~Wang and S.~Zhao.
\newblock Solution sets of finite fuzzy relation equations with sup–inf
  composition over bounded brouwerian lattices.
\newblock \emph{Information Sciences}, 234:\penalty0 80--85, 2013.

\bibitem[Perfilieva(2004)]{P.04}
I.~Perfilieva.
\newblock Fuzzy function as an approximate solution to a system of fuzzy
  relation equations.
\newblock \emph{Fuzzy Sets and Systems}, 147\penalty0 (3):\penalty0 363--383,
  2004.

\bibitem[Perfilieva(2013)]{P.13}
I.~Perfilieva.
\newblock Finitary solvability conditions for systems of fuzzy relation
  equations.
\newblock \emph{Information Sciences}, 234:\penalty0 29--43, 2013.

\bibitem[Perfilieva and Noskov{\' a}(2008)]{PN.08}
I.~Perfilieva and L.~Noskov{\' a}.
\newblock System of fuzzy relation equations with inf-→ composition: Complete
  set of solutions.
\newblock \emph{Fuzzy Sets and Systems}, 159\penalty0 (17):\penalty0
  2256--2271, 2008.

\bibitem[Perfilieva and Nov{\' a}k(2007)]{PN.07}
I.~Perfilieva and V.~Nov{\' a}k.
\newblock System of fuzzy relation equations as a continuous model of if–then
  rules.
\newblock \emph{Information Sciences}, 177\penalty0 (16):\penalty0 3218--3227,
  2007.

\bibitem[q.~Xiong and p.~Wang(2012)]{XW.12}
Q.~q.~Xiong and X.~p.~Wang.
\newblock Fuzzy relational equations on complete brouwerian lattices.
\newblock \emph{Information Sciences}, 193:\penalty0 141--152, 2012.

\bibitem[Qiao and Zhu(2020)]{QZ.20}
S.~Qiao and P.~Zhu.
\newblock Extremal solutions to fuzzy relation equations and inequalities with
  three unknowns.
\newblock \emph{Journal of Intelligent \& Fuzzy Systems}, 38:\penalty0
  5055–5076, 2020.

\bibitem[Qiu et~al.(2021)Qiu, Li, and Yang]{QLY.21}
J.~Qiu, G.~Li, and X.~Yang.
\newblock Arbitrary-term-absent max-product fuzzy relation inequalities and its
  lexicographic minimal solution.
\newblock \emph{Information Sciences}, 567:\penalty0 167--184, 2021.

\bibitem[Qu et~al.(2014)Qu, Wang, and Lei]{QWL.14}
X.-B. Qu, X.-P. Wang, and M.-H. Lei.
\newblock Conditions under which the solution sets of fuzzy relational
  equations over complete brouwerian lattices form lattices.
\newblock \emph{Fuzzy Sets and Systems}, 234:\penalty0 34--45, 2014.

\bibitem[Raftery(2007)]{R.07}
J.~Raftery.
\newblock Representable idempotent commutative residuated lattices.
\newblock \emph{Transactions of the American Mathematical Society},
  359\penalty0 (9):\penalty0 4405--4427, 2007.

\bibitem[Sanchez(1974)]{S.74}
E.~Sanchez.
\newblock \emph{Equations de relations floues}.
\newblock PhD thesis, Facult{\' e} de M{\' e}decine de Marseille, 1974.

\bibitem[Sanchez(1976)]{S.76}
E.~Sanchez.
\newblock Resolution of composite fuzzy relation equations.
\newblock \emph{Information and Control}, 30\penalty0 (1):\penalty0 38--48,
  1976.

\bibitem[Stamenkovi{\' c} et~al.(2014)Stamenkovi{\' c}, {\' C}iri{\' c}, and
  Ignjatovi{\' c}]{SCI.14}
A.~Stamenkovi{\' c}, M.~{\' C}iri{\' c}, and J.~Ignjatovi{\' c}.
\newblock Reduction of fuzzy automata by means of fuzzy quasi-orders.
\newblock \emph{Information Sciences}, 275:\penalty0 168 -- 198, 2014.

\bibitem[Stamenkovi{\'{c}} et~al.(2021)Stamenkovi{\'{c}}, {\'{C}}iri{\'{c}},
  and Djurdjanovi{\'{c}}]{SCD.21}
A.~Stamenkovi{\'{c}}, M.~{\'{C}}iri{\'{c}}, and D.~Djurdjanovi{\'{c}}.
\newblock Weakly linear systems for matrices over the max-plus quantale.
\newblock \emph{Discrete Event Dynamic Systems}, 2021.
\newblock \doi{10.1007/s10626-021-00342-4}.

\bibitem[Stankovi{\' c} et~al.(2017)Stankovi{\' c}, {\' C}iri{\' c}, and
  Ignjatovi{\' c}]{SCI.17}
I.~Stankovi{\' c}, M.~{\' C}iri{\' c}, and J.~Ignjatovi{\' c}.
\newblock Fuzzy relation equations and inequalities with two unknowns and their
  applications.
\newblock \emph{Fuzzy Sets and Systems}, 322:\penalty0 86 -- 105, 2017.

\bibitem[Stepanovi{\' c}(2018)]{S.18}
V.~Stepanovi{\' c}.
\newblock Fuzzy set inequations and equations with a meet-continuous codomain
  lattice.
\newblock \emph{Journal of Intelligent \& Fuzzy Systems}, 34:\penalty0
  4009--4021, 2018.

\bibitem[Sun(2012)]{S.12}
F.~Sun.
\newblock Conditions for the existence of the least solution and minimal
  solutions to fuzzy relation equations over complete brouwerian lattices.
\newblock \emph{Information Sciences}, 205:\penalty0 86–92, 2012.

\bibitem[Sun and b.~Qu(2021)]{SQ.21}
F.~Sun and X.~b.~Qu.
\newblock Resolution of fuzzy relation equations with increasing operations
  over complete lattices.
\newblock \emph{Information Sciences}, 570:\penalty0 451--467, 2021.

\bibitem[Sun et~al.(2020)Sun, b.~Qu, p.~Wang, and Zhu]{SQWZ.20}
F.~Sun, X.~b.~Qu, X.~p.~Wang, and L.~Zhu.
\newblock On pre-solution matrices of fuzzy relation equations over complete
  brouwerian lattices.
\newblock \emph{Fuzzy Sets and Systems}, 384:\penalty0 34--53, 2020.

\bibitem[Turunen(2020)]{T.20}
E.~Turunen.
\newblock Necessary and sufficient conditions for the existence of solution of
  generalized fuzzy relation equations {$A \Leftrightarrow X = B$}.
\newblock \emph{Information Sciences}, 536:\penalty0 351--357, 2020.

\bibitem[Wu et~al.(2021)Wu, Lur, Wen, and Lee]{WLWL.21}
Y.-K. Wu, Y.-Y. Lur, C.-F. Wen, and S.-J. Lee.
\newblock Analytical method for solving max-min inverse fuzzy relation.
\newblock \emph{Fuzzy Sets and Systems}, 2021.
\newblock \doi{https://doi.org/10.1016/j.fss.2021.08.019}.

\bibitem[Xiao et~al.(2019)Xiao, Zhu, Chen, and Yang]{XZCY.19}
G.~Xiao, T.~Zhu, Y.~Chen, and X.~Yang.
\newblock Linear searching method for solving approximate solution to system of
  max-min fuzzy relation equations with application in the instructional
  information resources allocation.
\newblock \emph{IEEE Access}, 7:\penalty0 65019--65028, 2019.

\bibitem[Yang(2014)]{Y.14}
S.-J. Yang.
\newblock An algorithm for minimizing a linear objective function subject to
  the fuzzy relation inequalities with addition–min composition.
\newblock \emph{Fuzzy Sets and Systems}, 255:\penalty0 41--51, 2014.

\bibitem[Yang(2018)]{Y.18}
S.-J. Yang.
\newblock Some results of the fuzzy relation inequalities with addition–min
  composition.
\newblock \emph{IEEE Transactions on Fuzzy Systems}, 26\penalty0 (1):\penalty0
  239--245, 2018.

\bibitem[Yang(2020{\natexlab{a}})]{Y.20}
X.~Yang.
\newblock Solutions and strong solutions of min-product fuzzy relation
  inequalities with application in supply chain.
\newblock \emph{Fuzzy Sets and Systems}, 384:\penalty0 54--74,
  2020{\natexlab{a}}.

\bibitem[Yang(2021{\natexlab{a}})]{Y.21.tfs}
X.~Yang.
\newblock Deviation analysis for the max-product fuzzy relation inequalities.
\newblock \emph{IEEE Transactions on Fuzzy Systems}, 29\penalty0 (12):\penalty0
  3782--3793, 2021{\natexlab{a}}.

\bibitem[Yang et~al.(2017)Yang, p.~Yang, and Hayat]{YYH.17}
X.-B. Yang, X.~p.~Yang, and K.~Hayat.
\newblock A new characterisation of the minimal solution set to max-min fuzzy
  relation inequalities.
\newblock \emph{Fuzzy Information and Engineering}, 9\penalty0 (4):\penalty0
  423--435, 2017.

\bibitem[Yang(2019)]{Y.19}
X.-P. Yang.
\newblock Evaluation model and approximate solution to inconsistent max-min
  fuzzy relation inequalities in p2p file sharing system.
\newblock \emph{Complexity}, 2019:\penalty0 1--11, 2019.

\bibitem[Yang(2020{\natexlab{b}})]{Y.20b}
X.-P. Yang.
\newblock Leximax minimum solution of addition-min fuzzy relation inequalities.
\newblock \emph{Information Sciences}, 524:\penalty0 184--198,
  2020{\natexlab{b}}.

\bibitem[Yang(2021{\natexlab{b}})]{Y.21}
X.-P. Yang.
\newblock Random-term-absent addition-min fuzzy relation inequalities and their
  lexicographic minimum solutions.
\newblock \emph{Fuzzy Sets and Systems}, 2021{\natexlab{b}}.

\bibitem[Yang et~al.(2015{\natexlab{a}})Yang, Zhou, and Cao]{YZC.15}
X.-P. Yang, X.-G. Zhou, and B.-Y. Cao.
\newblock Single-variable term semi-latticized fuzzy relation geometric
  programming with max-product operator.
\newblock \emph{Information Sciences}, 325:\penalty0 271--287,
  2015{\natexlab{a}}.

\bibitem[Yang et~al.(2015{\natexlab{b}})Yang, Zhou, and Cao]{YZC.15b}
X.-P. Yang, X.-G. Zhou, and B.-Y. Cao.
\newblock Multi-level linear programming subject to addition-min fuzzy relation
  inequalities with application in peer-to-peer file sharing system.
\newblock \emph{Journal of Intelligent \& Fuzzy Systems}, 28:\penalty0
  2679--2689, 2015{\natexlab{b}}.

\bibitem[Yang et~al.(2016)Yang, Zhou, and Cao]{YZC.16}
X.-P. Yang, X.-G. Zhou, and B.-Y. Cao.
\newblock Latticized linear programming subject to max-product fuzzy relation
  inequalities with application in wireless communication.
\newblock \emph{Information Sciences}, 358-359:\penalty0 44--55, 2016.

\bibitem[Yang et~al.(2018{\natexlab{a}})Yang, Lin, Zhou, and Cao]{YLZC.18}
X.-P. Yang, H.-T. Lin, X.-G. Zhou, and B.-Y. Cao.
\newblock Addition-min fuzzy relation inequalities with application in
  bittorrent-like peer-to-peer file sharing system.
\newblock \emph{Fuzzy Sets and Systems}, 343:\penalty0 126--140,
  2018{\natexlab{a}}.

\bibitem[Yang et~al.(2018{\natexlab{b}})Yang, Yuan, and Cao]{YYC.18}
X.-P. Yang, D.-H. Yuan, and B.-Y. Cao.
\newblock Lexicographic optimal solution of the multi-objective programming
  problem subject to max-product fuzzy relation inequalities.
\newblock \emph{Fuzzy Sets and Systems}, 341:\penalty0 92--112,
  2018{\natexlab{b}}.

\end{thebibliography}

\end{document}